\documentclass[runningheads]{llncs}

\usepackage{eccv}

\usepackage{eccvabbrv}

\usepackage{graphicx}
\usepackage{booktabs}

\usepackage[accsupp]{axessibility}  

\usepackage{colortbl}
\usepackage{wrapfig}
\usepackage[table]{xcolor}
\usepackage{array}
\usepackage{placeins}
\usepackage[percent]{overpic}
\usepackage{microtype}           
\usepackage{nicefrac}            
\usepackage{xspace}              
\usepackage{anyfontsize}
\usepackage{dsfont}              
\usepackage[ruled,vlined]{algorithm2e}

\usepackage{booktabs}            
\usepackage{nicematrix}         
\usepackage{multirow}
\usepackage{makecell}
\usepackage{bigstrut}
\usepackage{enumitem}            
\setitemize{label=\textbullet, leftmargin=*, nolistsep}
\usepackage{graphicx}
\usepackage{adjustbox}           
\usepackage{float}               
\usepackage{wrapfig}             
\usepackage{subcaption}          
\expandafter\def\csname ver@subfig.sty\endcsname{}  
\usepackage{fontawesome5}        
\usepackage{pifont}              
\definecolor{cvprblue}{rgb}{0.21,0.49,0.74}
\usepackage[pagebackref,breaklinks,colorlinks,allcolors=cvprblue]{hyperref}
\usepackage[all]{hypcap}         
\usepackage{url}                 
\usepackage[normalem]{ulem}      
\usepackage{lipsum}              
\usepackage{rotating}            
\usepackage{stackengine}         
\usepackage{caption}             
\usepackage{listings}            
\usepackage{soul}                
\usepackage{gradient-text}       
\usepackage{pgfplots}            
\pgfplotsset{compat=newest}
\usepackage{pgfplotstable}       
\usepackage{tikz}                
\usepackage{svg}                 

\usepackage{wrapstuff}

\definecolor{demphcolor}{RGB}{125,125,125}     

\newcommand{\xmark}{\textcolor{red}{\ding{55}}}     

\definecolor{cayenne}{RGB}{183, 55, 35}         
\definecolor{espresso}{RGB}{88, 41, 0}
\definecolor{IllinoisBlue}{HTML}{13294B}
\definecolor{IllinoisOrange}{HTML}{FF5F05}
\definecolor{SoftBeige}{HTML}{FEF5E7}
\definecolor{YaleBlue}{HTML}{2A5487}
\definecolor{CustomRed}{RGB}{233, 93, 34}      
\definecolor{CustomOrange}{RGB}{245, 132, 38}  
\definecolor{CustomPeach}{RGB}{255, 165, 82}   
\definecolor{CustomLightOrange}{RGB}{255, 204, 128} 
\definecolor{CustomYellow}{RGB}{255, 220, 148} 
\definecolor{CustomDarkGray}{RGB}{80, 80, 80}
\definecolor{CustomTeal}{HTML}{1BA1E2} 
\definecolor{BrightTeal}{RGB}{0, 166, 166}          
\definecolor{CustomMint}{RGB}{80, 201, 154}         
\definecolor{CustomPink}{RGB}{255, 130, 143}        
\definecolor{CustomPurple}{RGB}{136, 74, 178}       
\definecolor{CustomLightPurple}{RGB}{196, 151, 255} 
\definecolor{CustomLightLightPurple}{RGB}{230, 222, 255} 
\newcommand{\modelname}{DreamPartGen\xspace}
\newcommand{\modelnamenc}{DreamPartGen\xspace}
\newcommand{\modelnamegradient}
{\textbf{\gradientRGB{DreamPartGen}{27, 161, 226}{136, 74, 178}}\xspace}

\definecolor{IllinoisOrange}{HTML}{FF5F05}
\definecolor{IllinoisBlue}{HTML}{13294B}
\newcommand{\logoicon}{%
  \raisebox{-0.20\height}{\includegraphics[height=1.2em]{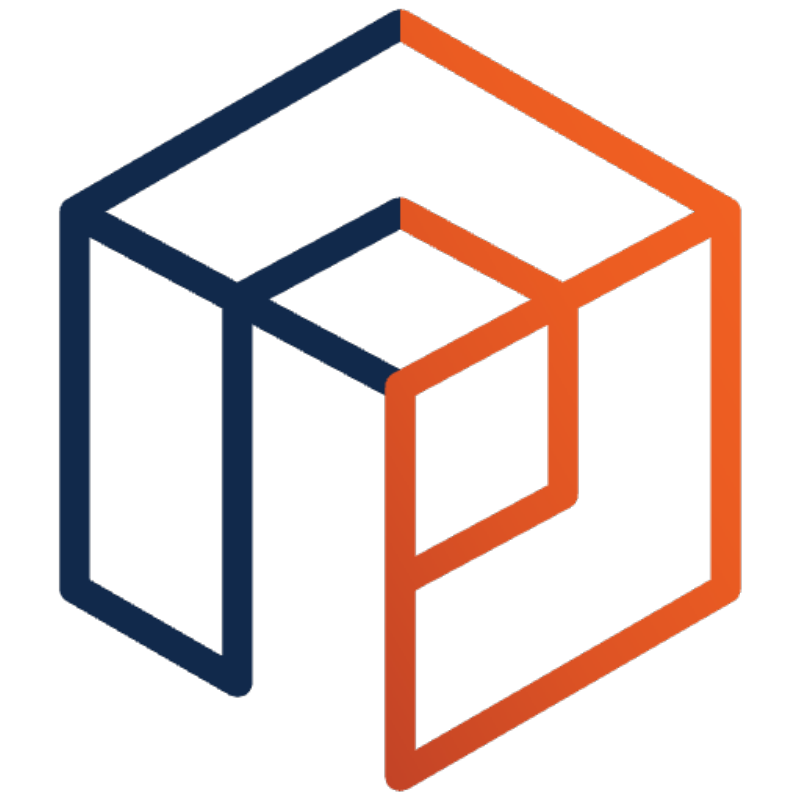}}%
}
\newcommand{\datasetnamegradient}
{\textbf{\gradientRGB{PartRel3D}{136, 74, 178}{27, 161, 226}}\xspace}

\newcommand{\datasetname}{PartRel3D\xspace}

\usepackage{hyperref}

\hypersetup{hypertexnames=false}
\usepackage{orcidlink}

\begin{document}

\title{\modelnamegradient: Semantically Grounded Part-Level 3D Generation via\\ Collaborative Latent Denoising} 

\titlerunning{\modelname}

\author{Tianjiao Yu\inst{1},
Xinzhuo Li\inst{1},
Muntasir Wahed\inst{1},
Jerry Xiong\inst{1},
Yifan Shen\inst{1}, 
Ying Shen\inst{1},
Ismini Lourentzou\inst{1} 
}

\authorrunning{T.~Yu et al.}

\institute{University of Illinois Urbana-Champaign\\
\email{\{ty41, lourent2\}@illinois.edu}}

\maketitle
\begin{center}
\vspace{-0.4cm}
\captionsetup{type=figure}\includegraphics[width=0.99\linewidth]{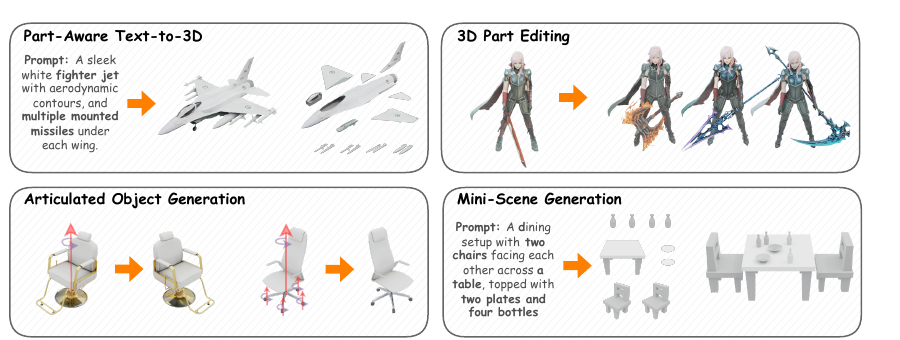}  
    \vspace{-0.3cm}
    \captionof{figure}{\modelnamegradient connects part-level geometry and appearance with language-driven relational semantics, providing precise control over how parts are modified, arranged, and contextualized. This unified representation enables a wide range of text-to-3D applications, including fine-grained part editing, articulated object generation, and mini-scene synthesis.}
    \label{fig:teaser}
    \vspace{-0.3cm}
\end{center}

\begin{abstract}
    Understanding and generating 3D objects as compositions of meaningful parts is fundamental to human perception and reasoning. However, most text-to-3D methods overlook the semantic and functional structure of parts. 
    While recent part-aware approaches introduce decomposition, they remain largely geometry-focused, lacking semantic grounding and failing to model how parts align with textual descriptions or their inter-part relations.
    We propose \modelnamegradient, a framework for semantically grounded, part-aware text-to-3D generation. \modelname introduces Duplex Part Latents (DPLs) that jointly model each part's geometry and appearance, and Relational Semantic Latents (RSLs) that capture inter-part dependencies derived from language.
    A synchronized co-denoising process enforces mutual geometric and semantic consistency, enabling coherent, interpretable, and text-aligned 3D synthesis. 
    Across multiple benchmarks, \modelnamenc delivers state-of-the-art performance in geometric fidelity ($\downarrow$60\% Chamfer Distance) and text–shape alignment ($\geq\uparrow$20\% CLIP/ULIP), while producing compositionally consistent and controllable parts.
    
    \noindent \logoicon~\href{https://plan-lab.github.io/dreampartgen}{\textcolor{IllinoisBlue}{PLAN Lab}~\textcolor{IllinoisOrange}{https://plan-lab.github.io/dreampartgen}}
    
\end{abstract}

\section{Introduction}
Many text prompts for 3D generation specify not only \emph{what} parts an object has, but \emph{how} they relate (\eg, a handle \emph{attached to} a mug, wheels \emph{symmetric} on a chassis, a lid \emph{on top of} a box). 
Capturing these part-level relations is crucial for controllable generation and downstream use cases such as part editing and articulated synthesis~\cite{mitra2014structure,yu2026part, laga2013geometry}. 
However, most text-to-3D methods operate on \emph{monolithic} latents that entangle geometry, appearance, and semantics, with no explicit representation of part identities or inter-part relations~\cite{poole2022dreamfusion, wang2023prolificdreamer, liang2024luciddreamer, lin2023magic3d, chen2023fantasia3d, liu2023zero, shi2023mvdream}. Recent part-aware methods take a step forward by synthesizing objects from part primitives guided by part segmentations or bounding boxes~\cite{liu2024part123, chen2025partgen, yang2025holopart, gao2024partgs, koo2023salad}. 
Although these approaches improve geometric granularity, they are still brittle to segmentation noise and can be difficult to scale across diverse categories and prompts. More importantly, many part-based frameworks still treat parts as geometrically isolated units. They do not model inter-part relations as explicit variables, and language remains largely non-operational. As a result, part-aware generation is not yet \emph{language-grounded assembly}.

We argue that true part-aware text-to-3D generation requires a different abstraction: a semantically grounded representation in which parts are meaningful entities, and language provides relational structure in addition to describing appearance. Concretely, we introduce \modelnamegradient, a language-grounded, collaborative part-latent diffusion framework that treats compositional semantics as an explicit representation during denoising. Each object is encoded into \textbf{\textcolor{CustomTeal}{Duplex Part Latents (DPLs)}}, paired 3D and 2D latent sequences that jointly capture a part’s geometry and appearance, whereas a learnable identifier embedding preserves slot identity across timesteps and instances, keeping parts trackable throughout diffusion. In parallel, we introduce \textbf{\textcolor{CustomPurple}{Relational Semantic Latents (RSLs)}}, compact text-derived latents that encode part-level attributes and inter-part relations. Rather than using language only as one-shot conditioning, \modelname performs synchronized co-denoising: DPLs and RSLs co-evolve through part-level and object-level synchronization so that geometry and appearance are refined under persistent, language-derived relational guidance, enforcing mutual geometric–semantic consistency.

To enable supervision at scale, we curate \datasetnamegradient, a large-scale relational dataset that augments each object with canonicalized functional and spatial triplets linking parts through explicit semantic predicates. These canonicalized relations are encoded into RSLs, allowing the model to learn assembly-level consistency directly from language. Trained on \datasetname, \modelname surpasses prior text-to-3D and part-aware baselines, achieving substantial improvements in geometric fidelity ($\downarrow$60\% CD, $\downarrow$41\% EMD) and text–shape alignment ($\geq$ $\uparrow$20\% CLIP/ULIP). We also evaluate generalization to rare parts and held-out relation predicates, improving over prior part-based baselines (\,$14.7\text{--}16.3\%\downarrow$ r-FID, $68.2\text{--}71.2\%\downarrow$ CD, $39.6\text{--}47.9\%\uparrow$ ULIP-T\,).
In summary, our contributions are:

\begin{itemize}
    \item We introduce \textbf{\modelname}, a language-grounded collaborative diffusion framework that unifies geometric, visual, and relational reasoning for coherent and interpretable part-level text-to-3D synthesis.
    \item We propose \textbf{DPLs} and \textbf{RSLs} as complementary representations that jointly encode part geometry, appearance, and inter-part relations and are refined together via synchronized co-denoising.
    \item We curate \textbf{\datasetname}, a large-scale relational dataset with 300K canonicalized functional and spatial triplets for explicit language-based supervision of inter-part relations across 175 object categories.
    \item \modelname achieves state-of-the-art performance across four benchmarks, improving geometric fidelity by $\downarrow$60\% CD / $\downarrow$41\% EMD and text–shape alignment by $\geq$ $\uparrow$20\% CLIP/ULIP, and showing the same representation supports relational part editing, articulated generation, and mini-scene synthesis.
\end{itemize}

\section{Related Work}
\noindent \textbf{Text-to-3D Generation.}  
Early text-to-3D approaches such as DreamFusion~\cite{poole2022dreamfusion}, ProlificDreamer~\cite{wang2023prolificdreamer}, and LucidDreamer~\cite{liang2024luciddreamer} leverage the idea of score distillation sampling (SDS) to generate 3D assets from 2D diffusion priors.  While effective for producing single objects, SDS approaches often suffer from low fidelity and poor multi-view consistency~\cite{shi2023mvdream, lin2023magic3d, chen2023fantasia3d, qiu2024richdreamer, li2023sweetdreamer, hu2024efficientdreamer}. Recent works improve training stability and geometric realism by incorporating differentiable rendering with explicit 3D representations, including Gaussian splatting in DreamGaussian~\cite{tang2024dreamgaussian} and GaussianDreamer~\cite{yi2024gaussiandreamer}, voxel- or mesh-based parameterizations in Clay~\cite{zhang2024clay}, and hybrid autoregressive architectures such as Trellis~\cite{xiang2025structured, yu2025core3d, lu2025unified}. These advances establish strong foundations for high-quality 3D generation, but typically focus on whole objects, without modeling explicit part structure or relational semantics. \looseness-1

\vspace{0.2cm}
\noindent \textbf{Part-level 3D Generation.}  
To address the limitations of object generation, several methods introduce part-aware modeling ~\cite{lin2026partcrafter,yang2025omnipart,hertz2022spaghetti, li2024pasta, yang2025holopart, ding2025fullpart, chen2026autopartgen}. Part123~\cite{liu2024part123} and Salad~\cite{koo2023salad} focus on part segmentation and assembly, while PartGen~\cite{chen2025partgen} leverages part decomposition for generative modeling. CoPart~\cite{dong2025one} extends diffusion models with dual priors over part-level 2D and 3D latents, enabling cross-modality and cross-part mutual guidance. 
Additionally, works such as PartGS~\cite{gao2024partgs} and Part$^2$GS~\cite{yu2026part} adapt Gaussian splatting for articulated part-aware generation, demonstrating that part supervision yields controllable and physically plausible synthesis~\cite{liu2025building, lu2025dreamart,shen2026gaussianart}. Despite these advances, prior part-aware generation methods rely heavily on geometric signals such as bounding boxes~\cite{zhu2025partsam}, leaving
language guidance underexplored, whereas a related line of work uses vision--language models for part-level {perception} rather than generation \cite{liu2023partslip, kareem2024paris3d, nguyen2025calico, wahed2024prima, li2025counterfactual}.
\modelname introduces explicit relational semantic signals that persist throughout denoising, providing both fine-grained part refinement and relation-aware global planning cues directly from natural language.

\begin{figure*}[t!]
    \centering
    \includegraphics[width=0.99\textwidth]{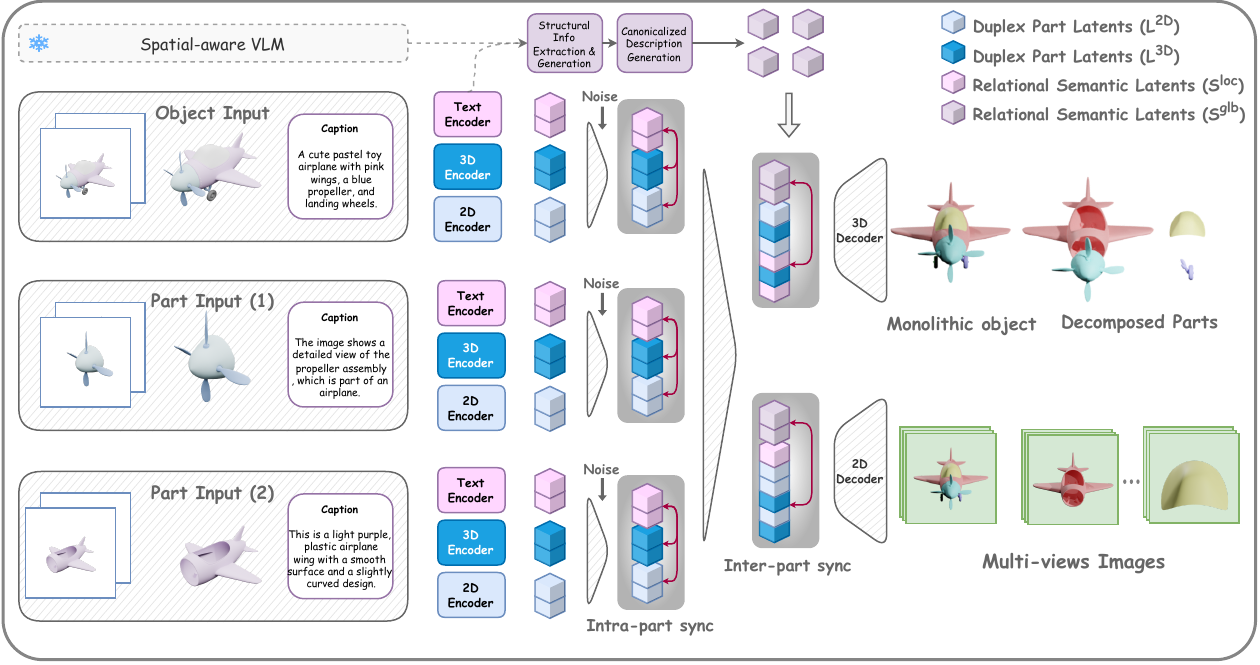}
    \vspace{-0.3cm}
    \caption{\textbf{\modelnamegradient Overview.} 
    \modelname performs text-guided 3D generation by jointly denoising geometry, appearance, and relational semantics. Each object is decomposed into parts represented as \textbf{\textcolor{CustomTeal}{Duplex Part Latents (DPLs)}} from 3D and 2D encoders, while \textbf{\textcolor{CustomPurple}{Relational Semantic Latents (RSLs)}} encode text-derived details and global structure. Through intra-part (geometry–appearance alignment) and inter-part (relational planning via language) synchronization, \modelnamenc co-denoises DPLs and RSLs to generate coherent, semantically grounded, part-aware 3D objects.}
    \label{fig:main_method}
    \vspace{-0.1cm}
\end{figure*}

\section{\modelnamegradient Method}\label{sec:method_main}
While recent part-level formulations improve local shape and texture modeling~\cite{chen2025partgen, chen2025ultra3d, yan2025x}, they primarily focus on representation quality and do not explicitly preserve \emph{text-derived semantics} throughout denoising, which limits their text-to-3D capability and fine-grained controllability. Our key novelty is to introduce \emph{persistent, language-derived relational semantic latents} that remain active throughout the denoising process, rather than using text only as a one-shot condition, and to synchronize them with part-level geometric latents. To this end, we formulate part-based 3D generation as a semantically grounded collaborative diffusion process between two complementary latent representations:
\textbf{\textcolor{CustomTeal}{\ding{182} Duplex Part Latents (DPLs)}} (Sec.~\ref{subsec:DPLs}), which encode geometry and appearance of individual parts in a modular and disentangled manner, and \textbf{\textcolor{CustomPurple}{\ding{183} Relational Semantic Latents (RSLs)}} (Sec.~\ref{subsec:RSLs}), a compact set of text-derived latent tokens that provide both local refinements and global planning signals. During denoising, DPLs and RSLs are synchronized through intra-part and inter-part attention (Sec.~\ref{subsec:sync}), enabling consistent part-level geometry-appearance alignment and language-guided part assembly. \Cref{fig:main_method} presents an overview of the method.

\subsection{\textcolor{CustomTeal}{Duplex Part Latents (DPLs)}}\label{subsec:DPLs}
The design of Duplex Part Latents (DPLs) is motivated by recent advances in structured latent representations for 3D generation, which demonstrate that compact latent sets can effectively encode both geometry and appearance ~\cite{xiang2025structured, yang2025omnipart, tang2026efficient}. However, existing unified latents primarily operate on voxel-aligned local features capturing shape and texture, but remain tied to spatial grids rather than semantic components. As a result, they lack modularity across objects and do not support explicit part-level disentanglement or relational reasoning. 
To address these limitations, we represent each object as a collection of $N$ semantic parts $O\!=\!\{p_i\}_{i=1}^N$, and encode each part using three complementary elements:
\begin{itemize}
    \item \textbf{3D tokens:} For each part mesh $p_i$, we sample surface points with associated normals and pass them through a 3D VAE encoder \cite{kingma2013auto, zhang20233dshape2vecset}, producing a latent sequence $\mathbf{L}^{\text{3D}}_i \in \mathbb{R}^{T_{\text{3D}} \times d}$, where $T_{\text{3D}}$ denotes the number of 3D latent tokens and $d$ their embedding dimension, capturing local geometry and spatial structure. 

\item \textbf{2D tokens:} Each part is also rendered from multiple viewpoints, and the resulting images are passed through a pretrained image VAE \cite{chen2024pixart}, yielding $\mathbf{L}^{\text{2D}}_i \in \mathbb{R}^{T_{\text{2D}} \times d}$, which encodes color, texture, and shading cues.

\item \textbf{Part-identity:}
To stabilize part tracking across denoising steps, we assign a learnable identifier embedding $e_i \in \mathbb{R}^d$ to each part. These identifiers act as persistent slot identities, binding each latent to its corresponding part and preventing slot swapping across denoising, while relational reasoning layers flexibly reorganize cross-part interactions.
\end{itemize}

\noindent Compared to prior structured latent designs \cite{lin2026partcrafter, xiang2025structured}, {Duplex Part Latents (DPLs)} are designed to preserve semantic independence while enabling language-conditioned relational reasoning. This yields several key benefits. First, the architecture is permutation-robust to the input ordering of parts, as the learnable part-identity embeddings prevent semantics from depending on the input part order. Second, the identifiers provide \emph{slot persistence} across denoising timesteps, improving stability of intra-part and inter-part synchronization. Third, because each part is represented as its own modular latent triplet $(\mathbf{L}^{\text{3D}}_i, \; \mathbf{L}^{\text{2D}}_i, e_i)$, DPLs naturally support cross-object generalization, enabling latent transfer between objects with shared functional components. Finally, DPLs are lightweight and modular, making them directly suitable for integration with the diffusion process, facilitating coherent multi-part synthesis and reasoning.\looseness-1

\subsection{\textcolor{CustomPurple}{Relational Semantic Latents (RSLs)}}
\label{subsec:RSLs}
While DPLs provide modular and disentangled representations for individual parts, they do not by themselves guarantee that the assembled object is globally coherent. This reflects a broader challenge in part-based 3D generation: local geometry and appearance can be faithfully synthesized, yet without explicit semantic coordination, the resulting object structure may violate plausible spatial or functional relations \cite{lin2023magic3d, wang2023prolificdreamer, chen2023fantasia3d}. To address this gap, we introduce {Relational Semantic Latents (RSLs)}, a compact set of \emph{language-derived latent tokens} that provide semantic control signals for part interactions through two roles: persistent global planners and diffused local refiners. In particular, global relational tokens $\mathbf{S}^{\text{glb}}$ persist as fixed structural conditions, while local semantic tokens $\mathbf{S}^{\text{loc}}$ are diffused and denoised alongside the part latents to refine part-level details.
\vspace{0.2cm}

\noindent \textbf{Global Relational Tokens.} 
At the object level, we extract relational phrases from whole-object and part-level descriptions (\eg, ``the seat is above the legs,'' ``the propeller is attached to the fuselage,'' ``the two wings are symmetric'', \etc). Each phrase is canonicalized into a triplet $(i,j,\rho)$, where $i$ and $j$ denote parts and $\rho$ is a relation predicate such as \texttt{support}, \texttt{attach}, \texttt{symmetry}, or \texttt{articulation}. These triplets are assembled into a relational graph and projected into the latent space (see Fig. \ref{fig:dataset_example}) to yield a set of global relational tokens:
\begin{equation}
\setlength{\abovedisplayskip}{7pt}
\setlength{\belowdisplayskip}{7pt}
\mathbf{S}^{\text{glb}} = \{ \mathbf{s}^{\text{glb}}_{ij,\rho} \}_{(i,j,\rho)\in\mathcal{R}}, \quad \mathbf{s}^{\text{glb}}_{ij,\rho} \in \mathbb{R}^d.    
\end{equation}
In this way, $\mathbf{S}^{\text{glb}}$ constitutes a relational graph latent, where each token encodes how two parts are semantically related. These tokens persist throughout the diffusion process and are injected into object-level synchronization, 
functioning both as semantic planners that specify inter-part relations and as structural conditions that enforce coherent assembly. 
Unlike prior geometry-based approaches, they are derived from natural language, embedding functional and structural priors without explicit geometric supervision.
\vspace{0.2cm}

\noindent \textbf{Local Semantic Tokens.} 
At the part level, we encode fine-grained semantic cues (\eg, ``metallic blade,'' ``wooden handle'', \etc) to refine material and appearance. Each phrase is encoded and projected into the latent space to yield $K_m$ local semantic tokens:\looseness-1
\begin{equation}
\setlength{\abovedisplayskip}{7pt}
\setlength{\belowdisplayskip}{7pt}
\mathbf{S}^{\text{loc}}  = \{ \mathbf{s}^{\text{loc}}_m \}_{m=1}^{K_m}, \quad \mathbf{s}^{\text{loc}}_m \in \mathbb{R}^d,  
\end{equation}
which directly interact with the structural DPL tokens to enhance geometric fidelity and appearance under semantic constraints.
Compared to geometry-only latents, RSLs are compact, interpretable, and flexible: their number adapts to object complexity, and additional tokens can be easily obtained by generating short textual descriptions for new parts or relations.
Unlike one-shot text conditioning  \cite{poole2022dreamfusion, lin2023magic3d}, we inject these tokens at every denoising step, enabling iterative semantic refinement. During diffusion, we apply the standard forward noising process to obtain $\mathbf{S}^{\text{loc},t}$ from the clean tokens $\mathbf{S}^{\text{loc}}$.
The noised local semantic tokens $\mathbf{S}^{\text{loc},t}$ are injected at each step $t$ to synchronize with the noised part latents and refine part-specific appearance.

\subsection{Semantically-Grounded Part Generation}
\label{subsec:sync}
We first instantiate DPLs by encoding each part mesh $p_i$ into geometry and appearance token sequences $(\mathbf{L}^{\text{3D}}_i,\mathbf{L}^{\text{2D}}_i)$ using the 3D VAE encoder and the pretrained image VAE encoder introduced in \Cref{subsec:DPLs}, and tag each part with a learnable identifier $e_i$.
We instantiate RSLs by encoding extracted relational/attribute phrases with a frozen text encoder $\mathcal{E}_{\text{text}}$ \cite{team2024gemma} followed by a learned projection $\phi_{\text{text}}$, yielding $(\mathbf{S}^{\text{glb}},\mathbf{S}^{\text{loc}})$.
To enable coherent generation, DPLs and RSLs interact throughout denoising via a two-level synchronization mechanism.
Specifically, we perform diffusion over the noised part latents $\{\mathbf{L}^{\text{3D},t}_i,\mathbf{L}^{\text{2D},t}_i\}_{i=1}^N$ and the noised local semantic tokens $\mathbf{S}^{\text{loc},t}$, while keeping the global relational tokens $\mathbf{S}^{\text{glb}}$ persistent as fixed structural conditions.
At each step $t$, we first apply \emph{intra-part synchronization} to align geometry and appearance within each part under local semantic guidance, and then apply \emph{inter-part synchronization} to propagate context across parts and enforce global relational constraints.

\vspace{0.3cm}
\noindent\textbf{Intra-Part Synchronization.}
At diffusion step $t$, each part $p_i$ is represented by a noised geometry latent sequence $\mathbf{L}^{\text{3D},t}_i$ and a noised appearance latent sequence $\mathbf{L}^{\text{2D},t}_i$.
We first synchronize these two streams to maintain intra-part geometry-appearance consistency, and then inject {noised} local semantic tokens $\mathbf{S}^{\text{loc},t}$ to refine part-specific geometric and visual details according to semantic cues. 
Formally,\looseness-1
\begin{equation}
\setlength{\abovedisplayskip}{7pt}
\setlength{\belowdisplayskip}{7pt}
\begin{aligned}
\mathbf{L}^{\text{3D},t}_i &\leftarrow \mathbf{L}^{\text{3D},t}_i + \alpha_{\text{3D}} \cdot \mathrm{Attn}(\mathbf{L}^{\text{3D},t}_i, \mathbf{L}^{\text{2D},t}_i), \\
\mathbf{L}^{\text{2D},t}_i &\leftarrow \mathbf{L}^{\text{2D},t}_i + \alpha_{\text{2D}} \cdot \mathrm{Attn}(\mathbf{L}^{\text{2D},t}_i, \mathbf{L}^{\text{3D},t}_i), \\
\mathbf{L}^{\text{3D},t}_i &\leftarrow \mathbf{L}^{\text{3D},t}_i + \lambda_{\text{3D}} \cdot \mathrm{Attn}(\mathbf{L}^{\text{3D},t}_i, \mathbf{S}^{\text{loc},t}), \\
\mathbf{L}^{\text{2D},t}_i &\leftarrow \mathbf{L}^{\text{2D},t}_i + \lambda_{\text{2D}} \cdot \mathrm{Attn}(\mathbf{L}^{\text{2D},t}_i, \mathbf{S}^{\text{loc},t}),
\end{aligned}
\end{equation}
where $\alpha_{\text{3D}}, \alpha_{\text{2D}}, \lambda_{\text{3D}}, \lambda_{\text{2D}}$ are learnable fusion coefficients. 

\noindent \textbf{Inter-Part Synchronization.}
After intra-part alignment, we propagate context across parts to encourage globally consistent assembly.
This is done in two complementary ways: (i) direct message passing among all part latents to share global context, and (ii) relational guidance from persistent global tokens $\mathbf{S}^{\text{glb}}$ that encode inter-part predicates (\eg, \texttt{support}, \texttt{attach}, \texttt{symmetry}, \texttt{articulation}).
Finally, we update $\mathbf{S}^{\text{glb}}$ via bottom-up grounding from the current part latents, refining the relational plan based on synthesized geometric and appearance evidence.
Concretely,
\begin{equation}
\setlength{\abovedisplayskip}{7pt}
\setlength{\belowdisplayskip}{7pt}
\begin{aligned}
&\mathbf{L}^{\text{3D},t}_i \leftarrow \mathbf{L}^{\text{3D},t}_i + \mathrm{Attn}\big(\mathbf{L}^{\text{3D},t}_i, \{\mathbf{L}^{\text{3D},t}_j\}_{j=1}^N\big), \\
&\mathbf{L}^{\text{2D},t}_i \leftarrow \mathbf{L}^{\text{2D},t}_i + \mathrm{Attn}\big(\mathbf{L}^{\text{2D},t}_i, \{\mathbf{L}^{\text{2D},t}_j\}_{j=1}^N\big), \\
&\mathbf{L}^{\text{3D},t}_i \leftarrow \mathbf{L}^{\text{3D},t}_i + \beta_{\text{3D}} \cdot \mathrm{Attn}(\mathbf{L}^{\text{3D},t}_i, \mathbf{S}^{\text{glb}}), \\
&\mathbf{L}^{\text{2D},t}_i \leftarrow \mathbf{L}^{\text{2D},t}_i + \beta_{\text{2D}} \cdot \mathrm{Attn}(\mathbf{L}^{\text{2D},t}_i, \mathbf{S}^{\text{glb}}), \\
&\mathbf{S}^{\text{glb}} \leftarrow \mathbf{S}^{\text{glb}} + \eta \cdot \mathrm{Attn}\Big(\mathbf{S}^{\text{glb}}, \{\mathrm{Pool}(\mathbf{L}^{\text{3D},t}_i, \mathbf{L}^{\text{2D},t}_i)\}_{i=1}^N\Big).
\end{aligned}
\end{equation}
where $\mathrm{Pool}(\cdot)$ aggregates each part's latent sequences into a compact summary for bottom-up grounding.
Note that $\mathbf{S}^{\text{glb}}$ is not a diffusion variable, but rather is updated deterministically as a planner state that remains available as a fixed relational condition at every timestep.

\noindent \textbf{Optimization.}
Training proceeds in two phases. In the first phase, we optimize diffusion objectives for both 3D and 2D DPLs under semantic conditioning from RSLs. For timesteps $t\!\sim\!\mathcal{U}\{1,\dots,T\}$ and noise $\varepsilon\!\sim\!\mathcal{N}(0,I)$, the per-part diffusion losses are:
\begin{equation}
\setlength{\abovedisplayskip}{7pt}
\setlength{\belowdisplayskip}{7pt}
\scalebox{0.89}{$
\begin{aligned}
\mathcal{L}_{\text{diff}}^{\text{3D}} &=
\frac{1}{N}\sum_{i=1}^N \mathbb{E}_{t,\varepsilon}\!
\left[\left\|\varepsilon - \mathcal{N}_{\text{3D}}~\!\big(\mathbf{L}^{\text{3D},t}_i,\mathbf{L}^{\text{2D},t}_i,\mathbf{S}^{\text{glb}},\mathbf{S}^{\text{loc},t},t\big)\right\|_2^2\right], \\
\mathcal{L}_{\text{diff}}^{\text{2D}} &=
\frac{1}{N}\sum_{i=1}^N \mathbb{E}_{t,\varepsilon}\!
\left[\left\|\varepsilon - \mathcal{N}_{\text{2D}}~\!\big(\mathbf{L}^{\text{2D},t}_i,\mathbf{L}^{\text{3D},t}_i,\mathbf{S}^{\text{glb}},\mathbf{S}^{\text{loc},t},t\big)\right\|_2^2\right].
\end{aligned}
$}
\end{equation}
In the second phase, we fine-tune the model jointly across the 3D and 2D part denoisers and synchronization modules, using an SNR-based weighting scheme.
The overall objective is
\begin{equation}
\setlength{\abovedisplayskip}{7pt}
\setlength{\belowdisplayskip}{7pt}
\mathcal{L} \;=\; \mathbb{E}_{t}\Big[
w_{\text{syn}}(t)\,\big(\mathcal{L}_{\text{diff}}^{\text{3D}}+\mathcal{L}_{\text{diff}}^{\text{2D}}\big)
\Big],
\end{equation}
where weights follow an SNR-based schedule $w_{\text{syn}}(t)\!=\!\frac{\mathrm{SNR}(t)}{1+\mathrm{SNR}(t)},
$
with $\mathrm{SNR}(t)\!=\!\alpha_t^2/\sigma_t^2$ defined by the diffusion coefficients.\looseness-1

\noindent \textbf{Inference.}
At test time, we encode the input prompt into local semantic tokens $\mathbf{S}^{\text{loc}}$ and global planner tokens $\mathbf{S}^{\text{glb}}$. When explicit triplets are available (\eg, provided by the user or an external parser), they are encoded as $\mathbf{S}^{\text{glb}}$; otherwise, we default to prompt-only conditioning instead of relying on external VLMs to supplement the triplets. This makes the VLM an optional semantic frontend for convenient relation extraction, rather than a required component of the generative backbone. We then initialize part latents by sampling Gaussian noise for the geometry and appearance streams, $\{\mathbf{L}^{\text{3D},T}_i,\mathbf{L}^{\text{2D},T}_i\}_{i=1}^N$, and initialize the local semantic stream by applying the same forward noising process used in training to obtain $\mathbf{S}^{\text{loc},T}$ from $\mathbf{S}^{\text{loc}}$. From timestep $T$ to $1$, we jointly denoise $\{\mathbf{L}^{\text{3D},t}_i,\mathbf{L}^{\text{2D},t}_i\}_{i=1}^N$ and $\mathbf{S}^{\text{loc},t}$ using the same part-level and object-level synchronization modules while conditioning on persistent $\mathbf{S}^{\text{glb}}$. After denoising, the final geometry latents $\{\mathbf{L}^{\text{3D},0}_i\}_{i=1}^N$ are decoded by the 3D VAE decoder to obtain part meshes, and the object is assembled from the decoded parts, with $\mathbf{L}^{\text{2D},0}_i$ used for appearance rendering when needed.\looseness-1

\section{\datasetnamegradient Dataset}
\label{subsec:dataset}

\begin{figure}[t!]
\centering
    \includegraphics[width=0.99\linewidth]{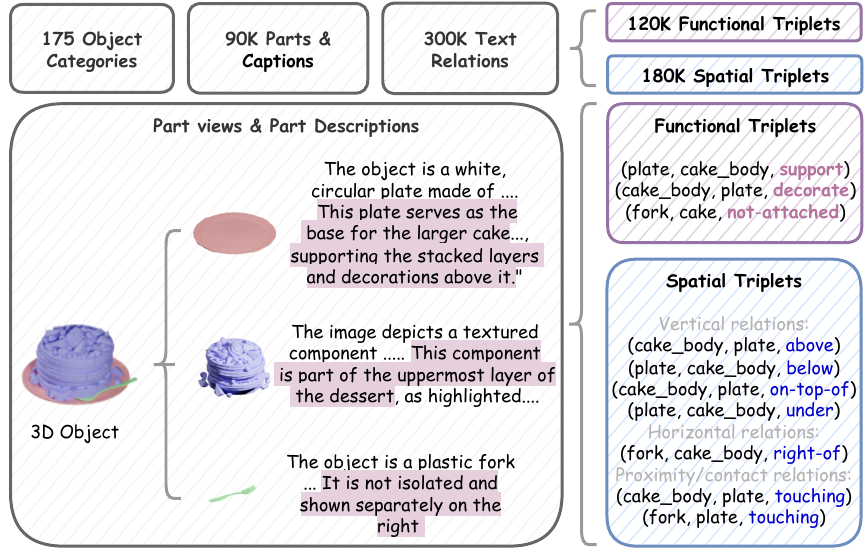}
    \vspace{-0.3cm}
    \caption{\textbf{\datasetname dataset overview} of structured functional and spatial triplets for fine-grained inter-part semantic supervision.}
    \label{fig:dataset_example}
\end{figure}

Existing 3D datasets such as PartNet~\cite{mo2019partnet},  Objaverse~\cite{deitke2023objaverse}, and PartVerse~\cite{dong2025one} provide large-scale geometric diversity but are limited in semantic grounding and relational coverage. They often include either geometry-only annotations or unconstrained text captions without consistent part correspondence, limiting their suitability for training models that understand object assembly (how parts connect) and semantics (what roles parts play).
To overcome these limitations, we introduce \datasetname, a large-scale, relationally annotated extension of PartVerse~\cite{dong2025one} that links part geometry, appearance, and language through explicit functional and spatial relationships. Each object in \datasetname is augmented with canonicalized triplets that encode both functional dependencies (\eg, support, attach, hinge) and spatial arrangements (\eg, above, touching, aligned-with), providing large-scale supervision of assembly-level semantics in 3D (\Cref{fig:dataset_example}).

\noindent \textbf{Functional Triplets} capture how parts interact in terms of support, attachment, and articulation.  
Given part- and object-level descriptions, we canonicalize phrases such as ``legs support seat'' or ``handle attached to body'' into triplets $(i,j,\rho^\text{function})$, where $i,j$ are part indices and $\rho^{\text{function}}$ is a functional predicate (\eg, \texttt{support}, \texttt{attach}, \texttt{hinge}, \texttt{symmetry}). \looseness-1

\noindent \textbf{Spatial Triplets} capture geometric and positional relations between parts. Each triplet has the same form $(i,j,\rho^\text{spatial})$, where $i$ and $j$ still index parts from the set of object parts $\mathcal{P}$, but $\rho^\text{spatial}$ is a predicate drawn from a controlled vocabulary of interpretable, assembly-relevant predicates. These include vertical relations (\texttt{above}, \texttt{below}, \texttt{on-top-of}, \texttt{under}), horizontal relations (\texttt{in-front-of}, \texttt{behind}, \texttt{left-of}, \texttt{right-of}), containment relations (\texttt{inside}, \texttt{surrounding}), symmetry/arrangement (\texttt{symmetric-with}, \texttt{parallel-to}, \texttt{aligned-with}), proximity and contact relations (\texttt{touching}, \texttt{attached-to}, \texttt{connected-with}).

The resulting \datasetname dataset contains approximately 11K part-labeled objects spanning 175 object categories, with over 90K individual parts and 300K canonicalized relational triplets. On average, each object contains 8.2 parts and 27 inter-part relations, providing dense structural supervision. Additional details, dataset statistics, and canonicalization criteria are available in the Appendix \ref{app:dataset}.

\section{Experiments}
\label{sec:experiments}

\textbf{Baselines.} We compare \modelnamenc{} against Trellis~\cite{xiang2025structured}, CLAY~\cite{zhang2024clay}, HoloPart ~\cite{yang2025holopart}, and PartCrafter~\cite{lin2026partcrafter}, as they represent the current state of the art in 3D generation. These methods collectively capture the diversity of contemporary approaches from structured latent representations  \cite{xiang2025structured} to explicit part generation and assembly part-aware text-driven 3D \cite{lin2026partcrafter}. Moreover, all of them provide open-source implementations, enabling fair and reproducible comparison under consistent training and evaluation protocols. Additional details in Appendix \ref{app:impldetails}.

\noindent \textbf{Metrics.} Following prior work~\cite{zhang2024clay}, we adopt both perceptual and structural metrics for text-to-3D evaluation. We report render-FID and render-KID (r-FID/r-KID) computed from multi-view renderings to assess visual fidelity, and P-FID/P-KID computed in 3D feature space using PointNet++~\cite{qi2017pointnet++}. Chamfer Distance (CD) and Earth Mover’s Distance (EMD) measure geometric precision. For text–shape alignment, we compute similarity with CLIP-ViT/L-14~\cite{radford2021learning} and ULIP~\cite{xue2023ulip}. In particular, ULIP-T is defined as the inner product between normalized ULIP embeddings of the caption $T$ and the generated shape $S$, $\text{ULIP-T}(T, S)=\langle E_T, E_S\rangle$, reflecting the semantic coherence between textual and geometric modalities. We further use the average pairwise Intersection-over-Union (IoU) to evaluate the geometric independence of generated part meshes. 
Specifically, we voxelize each generated part in a shared canonical space using a $64\times64\times64$ grid and compute the average pairwise IoU across all generated parts following \cite{lin2026partcrafter}. Lower IoU indicates less inter-part overlap and therefore better part disentanglement. The ideal case is that generated parts are non-intersecting while remaining composable into a plausible object consistent with the ground-truth structure.\looseness-1 

\begin{table*}[t]
    \centering
    \caption{\textbf{Quantitative evaluation on 3D object generation.} 
    Quantitative comparison with state-of-the-art methods on Objaverse, ShapeNet, ABO, and PartRel3D. 
    Highlighted \colorbox{CustomLightPurple}{best} and \colorbox{CustomLightLightPurple}{second-best} results.}
    \label{tab:quant-r3}
    \vspace{-0.3cm}
    \resizebox{0.99\linewidth}{!}{%
    \begin{tabular}{c c c c @{\hspace{4pt}} c c c @{\hspace{4pt}} c c c @{\hspace{4pt}} c c c}
        \toprule
         \multirow{2}{4em}{\textbf{Method}} & \multicolumn{3}{c}{\textbf{Objaverse}}  & \multicolumn{3}{c}{\textbf{ShapeNet}}  & \multicolumn{3}{c}{\textbf{ABO}} & \multicolumn{3}{c}{\textbf{PartRel3D}} \\ 
         \cmidrule(lr){2-4} \cmidrule(lr){5-7} \cmidrule(lr){8-10} \cmidrule(lr){11-13}
         \addlinespace[2pt]
        & CD$\downarrow$ & EMD$\downarrow$ & IoU$\downarrow$
        & CD$\downarrow$ & EMD$\downarrow$ & IoU$\downarrow$
        & CD$\downarrow$ & EMD$\downarrow$ & IoU$\downarrow$
        & CD$\downarrow$ & EMD$\downarrow$ & IoU$\downarrow$ \\
         \hline
        Trellis & 0.361 & 1.320 & - & 0.549 & 1.482 & - & 0.287 & 0.933 & - & 0.532 & 1.526 & - \\
        CLAY & 0.318 & 1.245 & - & 0.527 & 1.503 & - & 0.321 & 1.022 & - & 0.410 & 1.646 & - \\
        HoloPart & 0.334 & 1.298 & 0.494 & 0.478 & 1.354 & 0.542 & 0.269 & \cellcolor{CustomLightLightPurple}0.911 & 0.529 & \cellcolor{CustomLightLightPurple}0.355 & 1.623 & 0.716 \\
        PartCrafter & \cellcolor{CustomLightLightPurple}0.278 & \cellcolor{CustomLightLightPurple}1.107 & \cellcolor{CustomLightLightPurple}0.453 & \cellcolor{CustomLightLightPurple}0.451 & \cellcolor{CustomLightLightPurple}1.252 & \cellcolor{CustomLightPurple}0.499 & \cellcolor{CustomLightLightPurple}0.266 & 0.905 & \cellcolor{CustomLightLightPurple}0.505 & 0.371 & \cellcolor{CustomLightLightPurple}1.474 & \cellcolor{CustomLightLightPurple}0.700 \\
        \modelnamegradient & \cellcolor{CustomLightPurple}0.141 & \cellcolor{CustomLightPurple}0.810 & \cellcolor{CustomLightPurple}0.359 &
        \cellcolor{CustomLightPurple}0.222 & \cellcolor{CustomLightPurple}0.967 & \cellcolor{CustomLightLightPurple}0.503 &
        \cellcolor{CustomLightPurple}0.101 & \cellcolor{CustomLightPurple}0.531 & \cellcolor{CustomLightPurple}0.404 &
        \cellcolor{CustomLightPurple}0.081 & \cellcolor{CustomLightPurple}0.412 & \cellcolor{CustomLightPurple}0.304 \\

         \bottomrule
    \end{tabular}
    }
\end{table*}
\subsection{Quantitative Results}
We evaluate geometric reconstruction quality across Objaverse \cite{deitke2023objaverse}, ShapeNet \cite{chang2015shapenet}, ABO \cite{collins2022abo}, and PartRel3D. As shown in Table~\ref{tab:quant-r3}, \modelname consistently achieves the lowest CD and EMD on all benchmarks, outperforming prior methods by large margins ($\downarrow$60\% CD and $\downarrow$41\% EMD on average). 
Moreover, \modelname attains the lowest IoU scores ($\downarrow 27.2\%$ on average relative to the strongest baseline across each benchmark), reflecting stronger \textit{geometry independence}, \ie, the ability to generate non-intersecting yet composable parts that maintain object-level coherence.

\begin{wraptable}{r}{0.54\linewidth}
\vspace{-1.5cm}
\centering
\caption{\textbf{Text-shape alignment comparison.} Quantitative comparison on PartVerse. Highlighted \colorbox{CustomLightPurple}{best} and \colorbox{CustomLightLightPurple}{second-best}.}
\label{tab:quant-r2}
\setlength{\tabcolsep}{3pt}
    \resizebox{\linewidth}{!}{%
    \begin{tabular}{c c c c c}
        \toprule
        \textbf{Scope} & \textbf{Method} & CLIP(N-T)$\uparrow$ & CLIP(I-T)$\uparrow$ & ULIP-T$\uparrow$ \\
        \midrule
        \multirow{5}{*}{\textbf{Object-level}} 
        & Trellis      & 0.192 & 0.214 & 0.164 \\
        & CLAY         & \cellcolor{CustomLightLightPurple}0.194 & \cellcolor{CustomLightLightPurple}0.216 & 0.156 \\
        & HoloPart     & 0.186 & 0.206 & 0.155 \\
        & PartCrafter  & 0.187 & 0.207 & \cellcolor{CustomLightLightPurple}0.162 \\
        &  \modelnamegradient & \cellcolor{CustomLightPurple}0.235 & \cellcolor{CustomLightPurple}0.264 & \cellcolor{CustomLightPurple}0.197 \\
        \midrule
        \multirow{5}{*}{\textbf{Part-level}} 
        & Trellis      & 0.106 & 0.122 & 0.091 \\
        & CLAY         & 0.112 & 0.128 & \cellcolor{CustomLightLightPurple}0.096 \\
        & HoloPart     & 0.130 & \cellcolor{CustomLightLightPurple}0.141 & 0.113 \\
        & PartCrafter  & \cellcolor{CustomLightLightPurple}0.125 & 0.145 & 0.109 \\
        &  \modelnamegradient & \cellcolor{CustomLightPurple}0.179 & \cellcolor{CustomLightPurple}0.200 & \cellcolor{CustomLightPurple}0.153 \\
        \bottomrule
    \end{tabular}}
    \vspace{-0.7cm}
    \end{wraptable}
We further assess text–shape alignment on the PartVerse dataset following \cite{dong2025one}, where half of the test cases describe individual parts (\eg, “a chair leg”) and the rest correspond to complete objects. Table~\ref{tab:quant-r2} shows \modelname improves text-shape alignment over the strongest baseline across all metrics by
($\ge 20\%$) at the object level and
($\ge 35\%$) at the part level, highlighting the effectiveness of RSLs for fine-grained semantic grounding.

\begin{figure*}[t!]
    \centering
    \setlength{\tabcolsep}{16pt}
    \begin{tabular}{cccc}
        \textbf{Ground Truth} & \textbf{HoloPart} & \textbf{PartCrafter} & \modelnamegradient \\
    \end{tabular}
    \vspace{-4pt}
    \includegraphics[width=\linewidth]{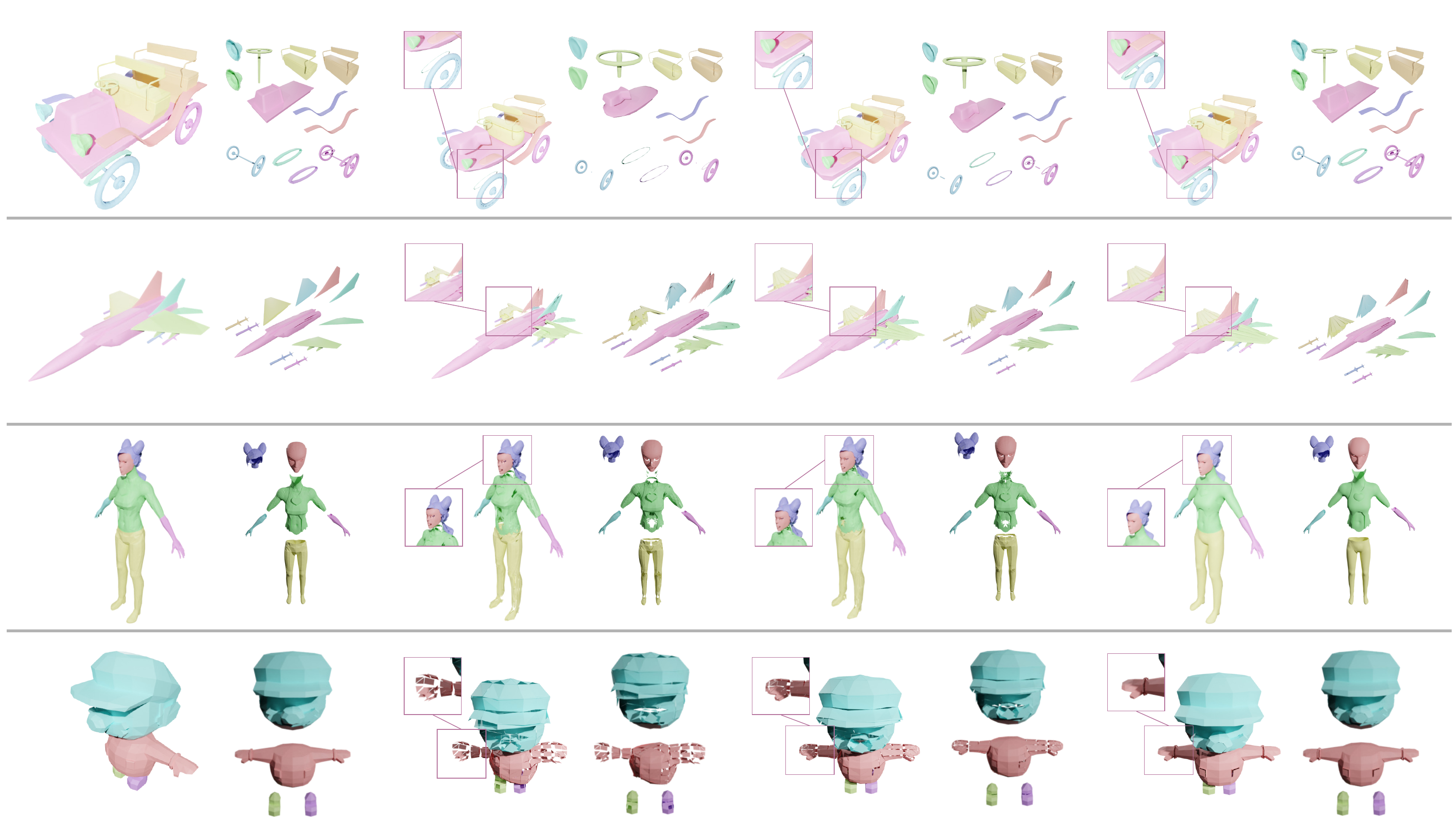}
    \vspace{-0.7cm}
    \caption{\textbf{Qualitative comparison on part-level 3D generation.} Across diverse object categories, \modelnamenc yields the most faithful decompositions, preserving clear part boundaries, correct topology, and consistent spatial alignment. Baselines frequently exhibit assembly failures such as missing or detached parts (\eg, wing/head), spatial drift of small components (\eg, wheels/mechanical parts floating off the chassis), and unstable attachments that create surface tearing or holes around high-contact regions (neck, torso, shoulders, limb joints).}
    \label{fig:qualitative_r1}
\end{figure*}

\subsection{Qualitative Results}

\Cref{fig:qualitative_r1} highlights that, across diverse object categories, \modelnamenc{} consistently generates 3D objects with part-consistent and physically plausible assemblies. Compared to the strongest baselines, HoloPart~\cite{yang2025holopart} and PartCrafter~\cite{lin2026partcrafter}, \modelnamenc{} preserves fine-grained geometry more faithfully, maintains inter-part relationships better, and respects global structural constraints that are frequently violated by prior approaches.
As illustrated, baselines frequently omit distorted parts or misplace them in space, for instance, generating wheels that float away from the chassis or misaligning small mechanical parts, leading to broken functional geometry in the first example. Similar failures appear in the second and third examples, where HoloPart produces a detached wing (airplane) or head (humanoid), and both baselines exhibit surface tearing and holes around the neck, torso, and shoulders, indicating incomplete and unstable attachment geometry. 
\modelnamenc{}, by contrast, generates watertight meshes with intact intra-part connections, smoother surfaces, and correctly integrated parts. Finally, in the last row, baselines suffer from hollow torsos, shredded hand geometry, and broken limb attachments, while \modelnamenc{} maintains coherent small-part geometry and avoids the severe tearing and disintegration observed in prior methods. Together, these results demonstrate that \modelnamenc{}'s relationally grounded generation maintains local part fidelity but also enforces globally consistent part connectivity even in complex articulated 3D structures.
\begin{figure}[t!]
    \centering
    \begin{minipage}[t]{0.48\textwidth}
        \vspace{0pt}
        \centering
        \includegraphics[width=0.99\linewidth]{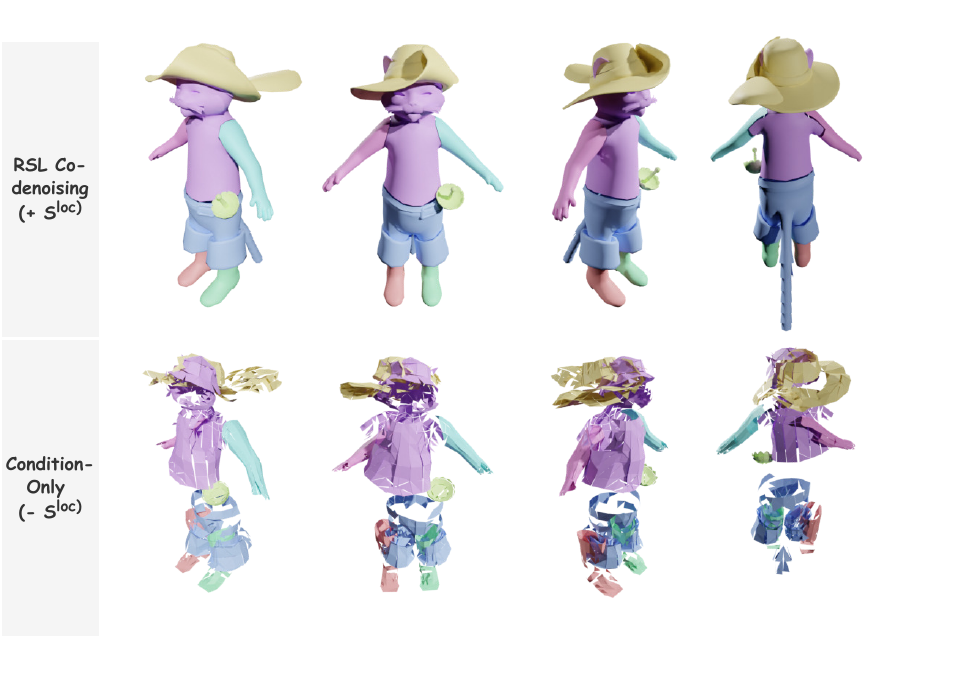}  
        \vspace{-0.9cm}
        \caption{\textbf{Ablation on the co-denoising process with local semantic tokens $\mathbf{S}^{\text{loc}}$.} The condition-only baseline ($- \mathbf{S}^{\text{loc}}$) yields coarse geometry and weak semantic coherence between parts.}
        \label{fig:ablate1}
    \end{minipage}\hfill
    \begin{minipage}[t]{0.48\textwidth}
        \vspace{0pt}
        \includegraphics[width=0.99\linewidth]{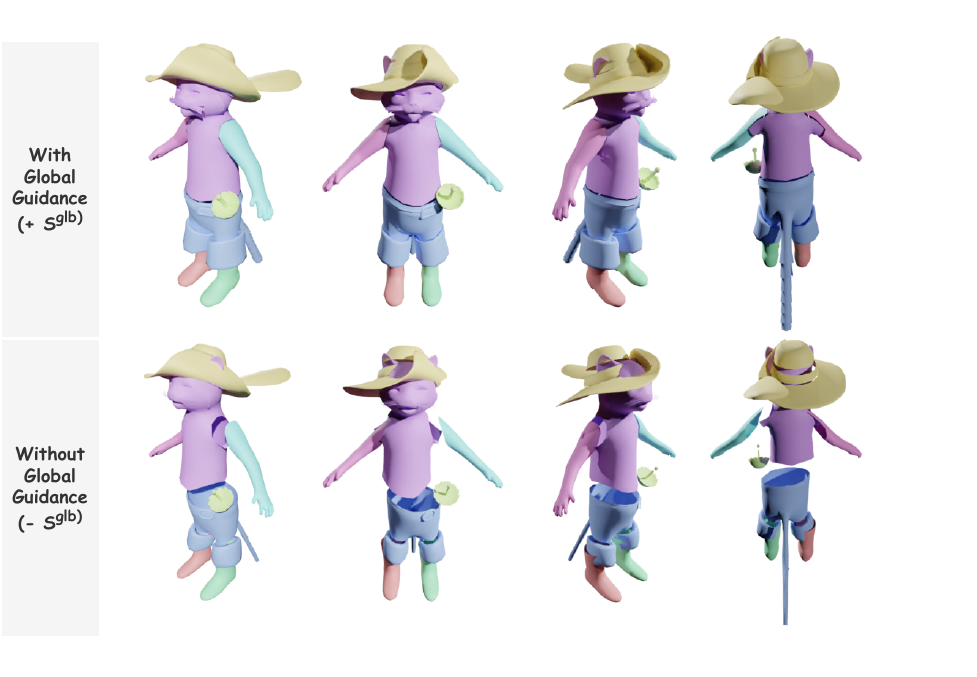}  
        \vspace{-0.9cm}
        \caption{\textbf{Ablation on the global relational tokens $\mathbf{S}^{\text{glb}}$.} The model exhibits part-level misalignment and spatial drift without $\mathbf{S}^{\text{glb}}$.}
        \label{fig:ablate2}
    \end{minipage}
\end{figure}

\subsection{Ablations}

\noindent\textbf{Relational Semantic Latents (RSLs) Qualitative Analysis.}
We qualitatively analyze the roles of the local semantic $\mathbf{S}^{\text{loc}}$ and global relational tokens $\mathbf{S}^{\text{glb}}$. 
For $\mathbf{S}^{\text{loc}}$, we compare against a conditioning-only baseline where text embeddings are injected only via timestep-wise cross-attention, without maintaining persistent semantic latents across denoising. As shown in \Cref{fig:ablate1}, conditioning-only yields coarser and less consistent surface geometry, with weaker semantic coherence between parts, indicating that co-denoising with $\mathbf{S}^{\text{loc}}$ is essential for high-fidelity part synthesis and stable semantic consistency throughout generation. 
For $\mathbf{S}^{\text{glb}}$, we remove the persistent global relational tokens and their object-level synchronization while keeping part-level denoising unchanged. Without global relational guidance, parts remain plausible in isolation but the assembled object exhibits\begin{wraptable}{r}{0.51\linewidth}
\vspace{-0.9cm}
\centering
\caption{\textbf{Ablation on Model Components.}
Evaluation on a fixed random \datasetname subset.
\colorbox{CustomLightPurple}{Best} and
\colorbox{CustomLightLightPurple}{second-best} results are highlighted.}
\label{tab:ablate-r1}
\setlength{\tabcolsep}{3pt}
\resizebox{\linewidth}{!}{%
\begin{tabular}{l c c c c}
    \toprule
    \textbf{Method} & CD$\downarrow$ & EMD$\downarrow$ & IoU$\downarrow$ & ULIP-T$\uparrow$ \\
    \midrule
    HoloPart
    & 0.226
    & 1.482
    & \cellcolor{CustomLightLightPurple}0.318
    & \cellcolor{CustomLightLightPurple}0.112 \\

    PartCrafter
    & \cellcolor{CustomLightLightPurple}0.219
    & \cellcolor{CustomLightLightPurple}1.403
    & 0.341
    & 0.101 \\
    \midrule

    \modelnamegradient\
    & \cellcolor{CustomLightPurple}0.145
    & \cellcolor{CustomLightPurple}0.771
    & \cellcolor{CustomLightPurple}0.212
    & \cellcolor{CustomLightPurple}0.158 \\

    \xmark\enspace $\mathbf{S}^{\text{glb}}$
    & 0.292 & 2.892 & 0.587 & 0.084 \\

    \xmark\enspace $\mathbf{S}^{\text{loc}}$
    & 0.781 & 5.764 & 0.652 & 0.089 \\

    \xmark\enspace \textbf{Part Identifier}
    & 0.277 & 1.709 & 0.438 & 0.091 \\
    \bottomrule
\end{tabular}}
\vspace{-0.9cm}
\end{wraptable}increased inter-part misalignment, weaker structural coherence, and spatial drift (\Cref{fig:ablate2}), confirming that global relational semantics are crucial for 
enforcing coherent object-level organization during denoising.

\vspace{0.2cm}
\noindent \textbf{RSLs and Part Identity Quantitative Analysis.} 
Table~\ref{tab:ablate-r1} summarizes the contribution of each component in \modelname, evaluated on a test subset of \datasetname dataset and compared against strong part-aware baselines. We assess performance using geometric fidelity (CD, EMD), part-level separation via pairwise IoU, and text–shape alignment through ULIP-T, resulting in a comprehensive view of both geometry and semantics. Removing the global relational tokens
(\xmark~$\mathbf{S}^{\mathrm{glb}}$) causes a major regression:
CD increases from 0.145 to 0.292 ($\uparrow$101.4\%),
EMD increases from 0.771 to 2.892 ($\uparrow$275.1\%), and part overlap rises from 0.212 to 0.587 ($\uparrow$176.9\%), while ULIP-T drops from 0.158 to 0.084 ($\downarrow$46.8\%), indicating that relational context is essential for preventing collisions and maintaining coherent assembly. Disabling the local semantic tokens (\xmark~$\mathbf{S}^{\text{loc}}$) further degrades performance: CD increases to $0.781$ ($\uparrow\!438.6\%$), EMD increases to $5.764$ ($\uparrow\!647.6\%$), IoU increases to $0.652$ ($\uparrow\!207.5\%$), and ULIP-T decreases to $0.089$ ($\downarrow\!43.7\%$), confirming the importance of jointly evolving part and semantic latents for stable generation.
Eliminating the part identifier module (\xmark~\textbf{Part Identifier}) also hurts disentanglement and semantics: IoU increases to $0.438$ ( $\uparrow\!106.6\%$), CD/EMD increase to $0.277/1.709$ ($\uparrow91.0\%/\uparrow121.7\%$), and ULIP-T drops to $0.091$ ($\downarrow\!42.4\%$), showing it helps preserve identity-consistent structure.

\vspace{0.2cm}
\noindent\textbf{Inference Efficiency.} \Cref{tab:inference_time_only} compares per-sample inference latency across representative 3D generation methods. Since these methods target different generation settings (object-level, part-level, and scene-level), we group comparisons by task type and interpret timings within each row. For \modelnamenc, we report the \emph{prompt-only} setting to isolate the cost of the generative backbone. Results show \modelnamenc remains efficient despite its semantic synchronization design.

\vspace{0.2cm}
\noindent\textbf{Additional Ablations.}  Appendix \ref{app:more_results} reports additional experiments, including perceptual metrics, part-level reconstruction quality, ablations on the number of RSL tokens, and robustness to relation parsing, rare parts, and held-out relations.

\begin{table}[t]
\centering
\caption{\textbf{Inference-time Comparison.} We report per-sample inference latency. Methods are grouped by task type (object-level, part-level, or scene-level generation); timings should be interpreted within each row. \colorbox{CustomLightPurple}{Best} is highlighted.}
\vspace{-0.4cm}
\setlength{\tabcolsep}{3pt}
\label{tab:inference_time_only}
\resizebox{\linewidth}{!}{
\begin{tabular}{l c c c c c c}
\toprule
\textbf{Task Setting} 
& \textbf{HoloPart} 
& \textbf{PartCrafter} 
& \textbf{TRELLIS} 
& \textbf{CLAY} 
& \textbf{MIDI} 
& \textbf{\modelnamegradient} \\
\midrule
Object Generation      & --   & --   & 95s  & 118s & --   & \colorbox{CustomLightPurple}{45s}   \\
Part-level Generation  & 21m  & 112s & --   & --   & --   & \colorbox{CustomLightPurple}{109s}  \\
3D Scene Generation    & --   & 64s  & --   & --   & 102s & \colorbox{CustomLightPurple}{52s}   \\
\bottomrule
\end{tabular}
}
\end{table} 

\subsection{Downstream Applications}

\noindent \textbf{Text-to-3D Scene Generation.}
\modelnamenc{} enables a wide range of part-aware 3D applications, including text-to-3D scene generation. In this task, the goal is to generate a coherent multi-object scene (a small scene) directly from a text prompt. 
During generation, each object is treated as a macro-part with aggregated DPLs, and a scene-level relational graph from canonicalized triplets $(o_i, o_j, \rho)$ encodes spatial and functional relations. Objects are first generated independently and then jointly refined in a brief synchronization step to produce the final, coherent scene. Additional details on the scene generation process can be found in Appendix \ref{app:applications}. 
As shown in \Cref{fig:result_miniscene},  \modelnamenc{} can synthesize multi-object scenes that respect part structure, spatial constraints, and global coherence. 
\modelnamenc{}'s DPLs assign persistent, semantically meaningful slots for every part category, ensuring that the model explicitly reasons over fine-grained sub-components (\eg, wooden chair legs) and part counts (\eg, four chairs).
Additional examples are provided in \Cref{fig:teaser} (bottom right) and in Appendix \ref{app:applications}.\looseness-1

\begin{figure}[!t]
    \centering
    \includegraphics[width=0.99\linewidth]{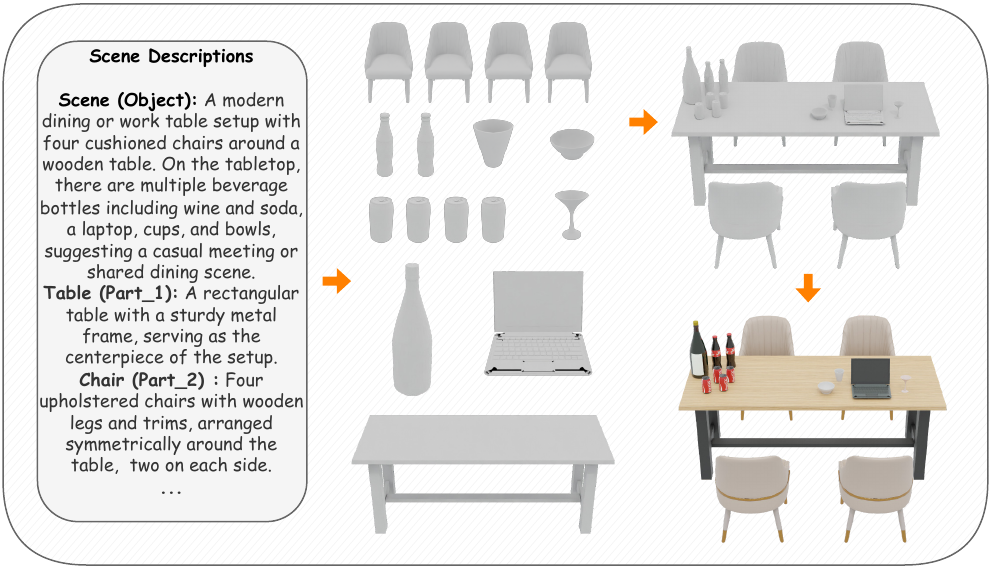}
    \vspace{-0.4cm}
    \caption{\textbf{Mini-scene generation}. Given a scene-level description, \modelnamenc{} can generate a complete, coherent 3D scene with physically plausible spatial layouts, capturing object geometry and fine-grained object- and part-level relations.}
    \label{fig:result_miniscene}
\end{figure}

\vspace{0.2cm}
\begin{wrapfigure}{r}{0.65\linewidth}
    \centering
    \vspace{-0.6cm}
    \includegraphics[width=0.99\linewidth]{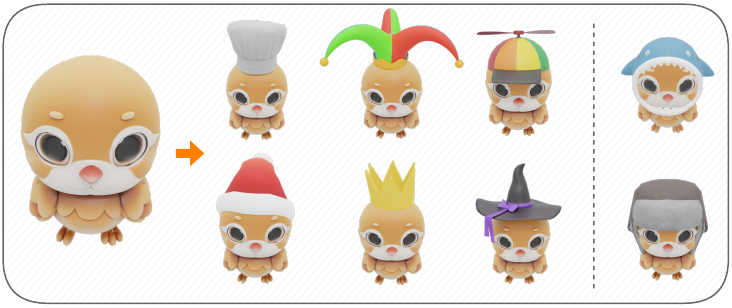}
    \vspace{-0.7cm}
    \caption{\textbf{Part-level editing.} From a source object (left), \modelnamenc{} can correctly execute relational edit prompts to place accessories on top of or around the head, while preserving geometry and spatial consistency.}
    \label{fig:result_part_editing}
    \vspace{-0.7cm}
\end{wrapfigure}\noindent \textbf{Text-to-3D Part Editing.} 
To edit a specific part, we isolate its DPLs and freeze all others while keeping the global relational context fixed. We then apply
localized re-denoising via partial DDIM inversion, optimizing only the target part’s latents, followed by a brief synchronization step to restore coherence with the full object. 
As illustrated in \Cref{fig:result_part_editing}, \modelnamenc{} accurately executes relational part editing prompts, producing clean, high-fidelity edits with seamless part-to-part coherence.
An additional editing example is shown in \Cref{fig:teaser} (top right). Details and additional qualitative examples can be found in Appendix \ref{app:applications}.

\section{Conclusion}
We introduce \modelname, a part-aware text-to-3D generation framework that bridges geometric structure and semantic reasoning through collaborative part-latent denoising. By coupling Duplex Part Latents (DPLs) with Relational Semantic Latents (RSLs), \modelnamenc{} jointly models geometry, appearance, and inter-part relations, enabling coherent, interpretable, and controllable 3D synthesis.
Beyond single-object generation, \modelname enables a broad suite of part-centric applications, including relational part editing and compositional scene generation, highlighting the benefits of explicitly modeling 3D objects through structured, semantically grounded part latents. An exciting future direction is to extend this formulation to densely entangled structures, such as bushes and other highly intricate objects or scenes.

\section*{Acknowledgments}
This research was partially supported by Google, the Google TPU Research Cloud (TRC) program, the Lambda Research Grant, the NSF CAREER Award \#2542328, and the U.S. Army under contract W5170125CA160. The views and conclusions contained herein are those of the authors and should not be interpreted as necessarily representing the official policies, either expressed or implied, of Google, Lambda, NSF, the U.S. Army, or the U.S. Government. The U.S. Government is authorized to reproduce and distribute reprints for governmental purposes notwithstanding any copyright annotation therein.

\bibliographystyle{splncs04}
\bibliography{main}

@String(CVPR  = {IEEE Conference on Computer Vision and Pattern Recognition (CVPR)})

@String(ICCV  = {International Conference on Computer Vision (ICCV)})

@String(ECCV  = {European Conference on Computer Vision (ECCV)})

@String(NeurIPS = {Advances in Neural Information Processing Systems (NeurIPS)})

@String(ICML  = {International Conference on Machine Learning (ICML)})

@String(ICLR  = {International Conference on Learning Representations (ICLR)})

@String(TOG   = {ACM Transactions on Graphics (TOG)})

@String(PMLR  = {Proceedings of Machine Learning Research (PMLR)})

@incollection{mitra2014structure,
  title={Structure-aware shape processing},
  author={Mitra, Niloy J and Wand, Michael and Zhang, Hao and Cohen-Or, Daniel and Kim, Vladimir and Huang, Qi-Xing},
  booktitle={ACM SIGGRAPH 2014 Courses},
  publisher={Association for Computing Machinery},
  pages={1--21},
  year={2014}
}

@inproceedings{kareem2024paris3d,
  title={Paris3d: Reasoning-based 3d part segmentation using large multimodal model},
  author={Kareem, Amrin and Lahoud, Jean and Cholakkal, Hisham},
  booktitle=ECCV,
  pages={466--482},
  year={2024},
  organization={Springer}
}

@inproceedings{liu2023partslip,
  title={Partslip: Low-shot part segmentation for 3d point clouds via pretrained image-language models},
  author={Liu, Minghua and Zhu, Yinhao and Cai, Hong and Han, Shizhong and Ling, Zhan and Porikli, Fatih and Su, Hao},
  booktitle=CVPR,
  pages={21736--21746},
  year={2023}
}

@article{laga2013geometry,
  title={Geometry and context for semantic correspondences and functionality recognition in man-made 3D shapes},
  author={Laga, Hamid and Mortara, Michela and Spagnuolo, Michela},
  journal={ACM Transactions on Graphics (TOG)},
  volume={32},
  number={5},
  pages={1--16},
  year={2013},
  publisher={ACM New York, NY, USA}
}

@article{poole2022dreamfusion,
  title={Dreamfusion: Text-to-3d using 2d diffusion},
  author={Poole, Ben and Jain, Ajay and Barron, Jonathan T and Mildenhall, Ben},
  journal={arXiv preprint arXiv:2209.14988},
  year={2022}
}

@article{wang2023prolificdreamer,
  title={Prolificdreamer: High-fidelity and diverse text-to-3d generation with variational score distillation},
  author={Wang, Zhengyi and Lu, Cheng and Wang, Yikai and Bao, Fan and Li, Chongxuan and Su, Hang and Zhu, Jun},
  journal=NeurIPS,
  pages={8406--8441},
  year={2023}
}

@inproceedings{liang2024luciddreamer,
  title={Luciddreamer: Towards high-fidelity text-to-3d generation via interval score matching},
  author={Liang, Yixun and Yang, Xin and Lin, Jiantao and Li, Haodong and Xu, Xiaogang and Chen, Yingcong},
  booktitle=CVPR,
  pages={6517--6526},
  year={2024}
}

@inproceedings{lin2023magic3d,
  title={Magic3d: High-resolution text-to-3d content creation},
  author={Lin, Chen-Hsuan and Gao, Jun and Tang, Luming and Takikawa, Towaki and Zeng, Xiaohui and Huang, Xun and Kreis, Karsten and Fidler, Sanja and Liu, Ming-Yu and Lin, Tsung-Yi},
  booktitle=CVPR,
  pages={300--309},
  year={2023}
}

@inproceedings{chen2023fantasia3d,
  title={Fantasia3d: Disentangling geometry and appearance for high-quality text-to-3d content creation},
  author={Chen, Rui and Chen, Yongwei and Jiao, Ningxin and Jia, Kui},
  booktitle=ICCV,
  pages={22246--22256},
  year={2023}
}

@inproceedings{liu2023zero,
  title={Zero-1-to-3: Zero-shot one image to 3d object},
  author={Liu, Ruoshi and Wu, Rundi and Van Hoorick, Basile and Tokmakov, Pavel and Zakharov, Sergey and Vondrick, Carl},
  booktitle=ICCV,
  pages={9298--9309},
  year={2023}
}

@article{shi2023mvdream,
  title={Mvdream: Multi-view diffusion for 3d generation},
  author={Shi, Yichun and Wang, Peng and Ye, Jianglong and Long, Mai and Li, Kejie and Yang, Xiao},
  journal={arXiv preprint arXiv:2308.16512},
  year={2023}
}

@inproceedings{liu2024part123,
  title={Part123: part-aware 3d reconstruction from a single-view image},
  author={Liu, Anran and Lin, Cheng and Liu, Yuan and Long, Xiaoxiao and Dou, Zhiyang and Guo, Hao-Xiang and Luo, Ping and Wang, Wenping},
  booktitle={ACM SIGGRAPH 2024 Conference Papers},
  pages={1--12},
  year={2024}
}

@inproceedings{chen2025partgen,
  title={Partgen: Part-level 3d generation and reconstruction with multi-view diffusion models},
  author={Chen, Minghao and Shapovalov, Roman and Laina, Iro and Monnier, Tom and Wang, Jianyuan and Novotny, David and Vedaldi, Andrea},
  booktitle=CVPR,
  pages={5881--5892},
  year={2025}
}

@article{yang2025holopart,
  title={Holopart: Generative 3d part amodal segmentation},
  author={Yang, Yunhan and Guo, Yuan-Chen and Huang, Yukun and Zou, Zi-Xin and Yu, Zhipeng and Li, Yangguang and Cao, Yan-Pei and Liu, Xihui},
  journal={arXiv preprint arXiv:2504.07943},
  year={2025}
}

@article{lin2026partcrafter,
  title={Partcrafter: Structured 3d mesh generation via compositional latent diffusion transformers},
  author={Lin, Yuchen and Lin, Chenguo and Pan, Panwang and Yan, Honglei and Yiqiang, Feng and Mu, Yadong and Fragkiadaki, Katerina},
  journal=NeurIPS,
  pages={35387--35415},
  year={2026}
}

@article{gao2024partgs,
  title={PartGS: Learning Part-aware 3D Representations by Fusing 2D Gaussians and Superquadrics},
  author={Gao, Zhirui and Yi, Renjiao and Huang, Yuhang and Chen, Wei and Zhu, Chenyang and Xu, Kai},
  journal={arXiv preprint arXiv:2408.10789},
  year={2024}
}

@article{chen2026autopartgen,
  title={AutoPartGen: Autoregressive 3D Part Generation and Discovery},
  author={Chen, Minghao and Wang, Jianyuan and Shapovalov, Roman and Monnier, Tom and Jung, Hyunyoung and Wang, Dilin and Ranjan, Rakesh and Laina, Iro and Vedaldi, Andrea},
  journal=NeurIPS,
  pages={153496--153521},
  year={2026}
}

@article{ding2025fullpart,
  title={FullPart: Generating each 3D Part at Full Resolution},
  author={Ding, Lihe and Dong, Shaocong and Li, Yaokun and Gao, Chenjian and Chen, Xiao and Han, Rui and Kuang, Yihao and Zhang, Hong and Huang, Bo and Huang, Zhanpeng and others},
  journal={arXiv preprint arXiv:2510.26140},
  year={2025}
}

@inproceedings{yu2026part,
  title={{Part$^2$GS}: Part-aware Modeling of Articulated Objects using 3D Gaussian Splatting},
  author={Yu, Tianjiao and Shah, Vedant and Wahed, Muntasir and Shen, Ying and Nguyen, Kiet A and Lourentzou, Ismini},
  booktitle=CVPR,
  pages={18913--18923},
  year={2026}
}

@inproceedings{koo2023salad,
  title={Salad: Part-level latent diffusion for 3d shape generation and manipulation},
  author={Koo, Juil and Yoo, Seungwoo and Nguyen, Minh Hieu and Sung, Minhyuk},
  booktitle=ICCV,
  pages={14441--14451},
  year={2023}
}

@article{lu2025unified,
  title={Unified Cross-Scale 3D Generation and Understanding via Autoregressive Modeling},
  author={Lu, Shuqi and Lin, Haowei and Yao, Lin and Gao, Zhifeng and Ji, Xiaohong and Liang, Yitao and Zhang, Linfeng and Ke, Guolin and others},
  journal={arXiv preprint arXiv:2503.16278},
  year={2025}
}

@article{yu2025core3d,
  title={{CoRe3D}: Collaborative Reasoning as a Foundation for 3D Intelligence},
  author={Yu, Tianjiao and Li, Xinzhuo and Shen, Yifan and Liu, Yuanzhe and Lourentzou, Ismini},
  journal={arXiv preprint arXiv:2512.12768},
  year={2025}
}

@article{yan2025x,
  title={X-part: high fidelity and structure coherent shape decomposition},
  author={Yan, Xinhao and Xu, Jiachen and Li, Yang and Ma, Changfeng and Yang, Yunhan and Wang, Chunshi and Zhao, Zibo and Lai, Zeqiang and Zhao, Yunfei and Chen, Zhuo and others},
  journal={arXiv preprint arXiv:2509.08643},
  year={2025}
}

@article{chen2025ultra3d,
  title={Ultra3d: Efficient and high-fidelity 3d generation with part attention},
  author={Chen, Yiwen and Li, Zhihao and Wang, Yikai and Zhang, Hu and Li, Qin and Zhang, Chi and Lin, Guosheng},
  journal={arXiv preprint arXiv:2507.17745},
  year={2025}
}

@article{wahed2024prima,
  title={Prima: Multi-image vision-language models for reasoning segmentation},
  author={Wahed, Muntasir and Nguyen, Kiet A and Juvekar, Adheesh Sunil and Li, Xinzhuo and Zhou, Xiaona and Shah, Vedant and Yu, Tianjiao and Yanardag, Pinar and Lourentzou, Ismini},
  journal={arXiv preprint arXiv:2412.15209},
  year={2024}
}

@article{zhu2025partsam,
  title={Partsam: A scalable promptable part segmentation model trained on native 3d data},
  author={Zhu, Zhe and Wan, Le and Xu, Rui and Zhang, Yiheng and Chen, Honghua and Dou, Zhiyang and Lin, Cheng and Liu, Yuan and Wei, Mingqiang},
  journal={arXiv preprint arXiv:2509.21965},
  year={2025}
}

@inproceedings{li2025counterfactual,
  title={Counterfactual segmentation reasoning: Diagnosing and mitigating pixel-grounding hallucination},
  author={Li, Xinzhuo and Juvekar, Adheesh and Zhang, Jiaxun and Liu, Xingyou and Wahed, Muntasir and Nguyen, Kiet A and Shen, Yifan and Yu, Tianjiao and Lourentzou, Ismini},
  booktitle=CVPR,
  pages={7450--7460},
  year={2026}
}

@inproceedings{xiang2025structured,
  title={Structured 3d latents for scalable and versatile 3d generation},
  author={Xiang, Jianfeng and Lv, Zelong and Xu, Sicheng and Deng, Yu and Wang, Ruicheng and Zhang, Bowen and Chen, Dong and Tong, Xin and Yang, Jiaolong},
  booktitle=CVPR,
  pages={21469--21480},
  year={2025}
}

@inproceedings{yang2025omnipart,
  title={Omnipart: Part-aware 3d generation with semantic decoupling and structural cohesion},
  author={Yang, Yunhan and Zhou, Yufan and Guo, Yuan-Chen and Zou, Zi-Xin and Huang, Yukun and Liu, Ying-Tian and Xu, Hao and Liang, Ding and Cao, Yan-Pei and Liu, Xihui},
  booktitle={Proceedings of the SIGGRAPH Asia 2025 Conference Papers},
  pages={1--12},
  year={2025}
}

@article{tang2026efficient,
  title={Efficient part-level 3d object generation via dual volume packing},
  author={Tang, Jiaxiang and Lu, Ruijie and Li, Max and Hao, Zekun and Li, Xuan and Wei, Fangyin and Song, Shuran and Zeng, Gang and Liu, Ming-Yu and Lin, Tsung-Yi},
  journal=NeurIPS,
  pages={27115--27137},
  year={2026}
}

@inproceedings{dong2025one,
  title={From one to more: Contextual part latents for 3d generation},
  author={Dong, Shaocong and Ding, Lihe and Chen, Xiao and Li, Yaokun and Wang, Yuxin and Wang, Yucheng and Wang, Qi and Kim, Jaehyeok and Gao, Chenjian and Huang, Zhanpeng and others},
  booktitle=CVPR,
  pages={8230--8240},
  year={2025}
}

@inproceedings{mo2019partnet,
  title={Partnet: A large-scale benchmark for fine-grained and hierarchical part-level 3d object understanding},
  author={Mo, Kaichun and Zhu, Shilin and Chang, Angel X and Yi, Li and Tripathi, Subarna and Guibas, Leonidas J and Su, Hao},
  booktitle=CVPR,
  pages={909--918},
  year={2019}
}

@article{zhang20233dshape2vecset,
  title={3dshape2vecset: A 3d shape representation for neural fields and generative diffusion models},
  author={Zhang, Biao and Tang, Jiapeng and Niessner, Matthias and Wonka, Peter},
  journal={ACM Transactions On Graphics (TOG)},
  volume={42},
  number={4},
  pages={1--16},
  year={2023},
  publisher={ACM New York, NY, USA}
}

@article{kingma2013auto,
  title={Auto-encoding variational bayes},
  author={Kingma, Diederik P and Welling, Max},
  journal={arXiv preprint arXiv:1312.6114},
  year={2013}
}

@inproceedings{chen2024pixart,
  title={Pixart-$\alpha$: Fast training of diffusion transformer for photorealistic text-to-image synthesis},
  author={Chen, Junsong and Yu, Jincheng and Ge, Chongjian and Yao, Lewei and Xie, Enze and Wang, Zhongdao and Kwok, James and Luo, Ping and Lu, Huchuan and Li, Zhenguo},
  booktitle=ICLR,
  volume={2024},
  pages={57611--57640},
  year={2024}
}

@inproceedings{yi2024gaussiandreamer,
  title={Gaussiandreamer: Fast generation from text to 3d gaussians by bridging 2d and 3d diffusion models},
  author={Yi, Taoran and Fang, Jiemin and Wang, Junjie and Wu, Guanjun and Xie, Lingxi and Zhang, Xiaopeng and Liu, Wenyu and Tian, Qi and Wang, Xinggang},
  booktitle=CVPR,
  pages={6796--6807},
  year={2024}
}

@article{li2024pasta,
  title={PASTA: Controllable Part-Aware Shape Generation with Autoregressive Transformers},
  author={Li, Songlin and Paschalidou, Despoina and Guibas, Leonidas},
  journal={arXiv preprint arXiv:2407.13677},
  year={2024}
}

@article{hertz2022spaghetti,
  title={Spaghetti: Editing implicit shapes through part aware generation},
  author={Hertz, Amir and Perel, Or and Giryes, Raja and Sorkine-Hornung, Olga and Cohen-Or, Daniel},
  journal={ACM Transactions on Graphics (TOG)},
  volume={41},
  number={4},
  pages={1--20},
  year={2022},
  publisher={ACM New York, NY, USA}
}

@inproceedings{hu2024efficientdreamer,
  title={Efficientdreamer: High-fidelity and robust 3d creation via orthogonal-view diffusion priors},
  author={Hu, Zhipeng and Zhao, Minda and Zhao, Chaoyi and Liang, Xinyue and Li, Lincheng and Zhao, Zeng and Fan, Changjie and Zhou, Xiaowei and Yu, Xin},
  booktitle=CVPR,
  pages={4949--4958},
  year={2024}
}

@article{li2023sweetdreamer,
  title={Sweetdreamer: Aligning geometric priors in 2d diffusion for consistent text-to-3d},
  author={Li, Weiyu and Chen, Rui and Chen, Xuelin and Tan, Ping},
  journal={arXiv preprint arXiv:2310.02596},
  year={2023}
}

@inproceedings{qiu2024richdreamer,
  title={Richdreamer: A generalizable normal-depth diffusion model for detail richness in text-to-3d},
  author={Qiu, Lingteng and Chen, Guanying and Gu, Xiaodong and Zuo, Qi and Xu, Mutian and Wu, Yushuang and Yuan, Weihao and Dong, Zilong and Bo, Liefeng and Han, Xiaoguang},
  booktitle=CVPR,
  pages={9914--9925},
  year={2024}
}

@article{zhang2024clay,
  title={Clay: A controllable large-scale generative model for creating high-quality 3d assets},
  author={Zhang, Longwen and Wang, Ziyu and Zhang, Qixuan and Qiu, Qiwei and Pang, Anqi and Jiang, Haoran and Yang, Wei and Xu, Lan and Yu, Jingyi},
  journal={ACM Transactions on Graphics (TOG)},
  volume={43},
  number={4},
  pages={1--20},
  year={2024},
  publisher={ACM New York, NY, USA}
}

@inproceedings{tang2024dreamgaussian,
  title={Dreamgaussian: Generative gaussian splatting for efficient 3d content creation},
  author={Tang, Jiaxiang and Ren, Jiawei and Zhou, Hang and Liu, Ziwei and Zeng, Gang},
  booktitle={International Conference on Learning Representations},
  volume={2024},
  pages={33879--33896},
  year={2024}
}

@article{lu2025dreamart,
  title={Dreamart: Generating interactable articulated objects from a single image},
  author={Lu, Ruijie and Liu, Yu and Tang, Jiaxiang and Ni, Junfeng and Wang, Yuxiang and Wan, Diwen and Zeng, Gang and Chen, Yixin and Huang, Siyuan},
  journal={arXiv preprint arXiv:2507.05763},
  year={2025}
}

@inproceedings{shen2026gaussianart,
  title={Gaussianart: Unified modeling of geometry and motion for articulated objects},
  author={Shen, Licheng and Zhang, Saining and Li, Honghan and Yang, Peilin and Huang, Zihao and Zhang, Zongzheng and Zhao, Hao},
  booktitle={2026 International Conference on 3D Vision (3DV)},
  pages={384--395},
  year={2026},
  organization={IEEE}
}

@inproceedings{liu2025building,
  title={Building interactable replicas of complex articulated objects via gaussian splatting},
  author={Liu, Yu and Jia, Baoxiong and Lu, Ruijie and Ni, Junfeng and Zhu, Song-Chun and Huang, Siyuan},
  booktitle={The Thirteenth International Conference on Learning Representations},
  year={2025}
}

@inproceedings{huang2025midi,
  title={Midi: Multi-instance diffusion for single image to 3d scene generation},
  author={Huang, Zehuan and Guo, Yuan-Chen and An, Xingqiao and Yang, Yunhan and Li, Yangguang and Zou, Zi-Xin and Liang, Ding and Liu, Xihui and Cao, Yan-Pei and Sheng, Lu},
  booktitle=CVPR,
  pages={23646--23657},
  year={2025}
}

@inproceedings{nguyen2025calico,
  title={CALICO: Part-Focused Semantic Co-Segmentation with Large Vision-Language Models},
  author={Nguyen, Kiet A and Juvekar, Adheesh and Yu, Tianjiao and Wahed, Muntasir and Lourentzou, Ismini},
  booktitle=CVPR,
  pages={4550--4561},
  year={2025}
}

@article{team2024gemma,
  title={Gemma 2: Improving open language models at a practical size},
  author={Team, Gemma and Riviere, Morgane and Pathak, Shreya and Sessa, Pier Giuseppe and Hardin, Cassidy and Bhupatiraju, Surya and Hussenot, L{\'e}onard and Mesnard, Thomas and Shahriari, Bobak and Ram{\'e}, Alexandre and others},
  journal={arXiv preprint arXiv:2408.00118},
  year={2024}
}

@inproceedings{collins2022abo,
  title={Abo: Dataset and benchmarks for real-world 3d object understanding},
  author={Collins, Jasmine and Goel, Shubham and Deng, Kenan and Luthra, Achleshwar and Xu, Leon and Gundogdu, Erhan and Zhang, Xi and Vicente, Tomas F Yago and Dideriksen, Thomas and Arora, Himanshu and others},
  booktitle=CVPR,
  pages={21126--21136},
  year={2022}
}

@article{chang2015shapenet,
  title={Shapenet: An information-rich 3d model repository},
  author={Chang, Angel X and Funkhouser, Thomas and Guibas, Leonidas and Hanrahan, Pat and Huang, Qixing and Li, Zimo and Savarese, Silvio and Savva, Manolis and Song, Shuran and Su, Hao and others},
  journal={arXiv preprint arXiv:1512.03012},
  year={2015}
}

@inproceedings{deitke2023objaverse,
  title={Objaverse: A universe of annotated 3d objects},
  author={Deitke, Matt and Schwenk, Dustin and Salvador, Jordi and Weihs, Luca and Michel, Oscar and VanderBilt, Eli and Schmidt, Ludwig and Ehsani, Kiana and Kembhavi, Aniruddha and Farhadi, Ali},
  booktitle=CVPR,
  pages={13142--13153},
  year={2023}
}

@article{bai2025qwen2,
  title={Qwen2.5-vl technical report},
  author={Bai, Shuai and Chen, Keqin and Liu, Xuejing and Wang, Jialin and Ge, Wenbin and Song, Sibo and Dang, Kai and Wang, Peng and Wang, Shijie and Tang, Jun and others},
  journal={arXiv preprint arXiv:2502.13923},
  year={2025}
}

@article{qi2017pointnet++,
  title={Pointnet++: Deep hierarchical feature learning on point sets in a metric space},
  author={Qi, Charles Ruizhongtai and Yi, Li and Su, Hao and Guibas, Leonidas J},
  journal=NeurIPS,
  year={2017}
}

@inproceedings{radford2021learning,
  title={Learning transferable visual models from natural language supervision},
  author={Radford, Alec and Kim, Jong Wook and Hallacy, Chris and Ramesh, Aditya and Goh, Gabriel and Agarwal, Sandhini and Sastry, Girish and Askell, Amanda and Mishkin, Pamela and Clark, Jack and others},
  booktitle=ICML,
  pages={8748--8763},
  year={2021},
  organization={PMLR}
}

@inproceedings{xue2023ulip,
  title={Ulip: Learning a unified representation of language, images, and point clouds for 3d understanding},
  author={Xue, Le and Gao, Mingfei and Xing, Chen and Mart{\'\i}n-Mart{\'\i}n, Roberto and Wu, Jiajun and Xiong, Caiming and Xu, Ran and Niebles, Juan Carlos and Savarese, Silvio},
  booktitle=CVPR,
  pages={1179--1189},
  year={2023}
}

\appendix
\clearpage
\setcounter{page}{1}


\vspace{-0.5cm}

\section{\datasetname Dataset}\label{app:dataset}
\noindent\textbf{Canonicalization.}  
When available, functional metadata is directly converted into triplets; otherwise, relations are generated using a pretrained VLM~\cite{bai2025qwen2} prompted with rendered views and part captions. Free-form relational phrases from captions or VLM outputs are normalized into a predicate vocabulary $\mathcal{V}_{\text{spatial}}$ through a two-step process:  
(i) \emph{Parsing}, where relational clauses are extracted from text (\eg, ``the seat is positioned right above the legs''), and  
(ii) \emph{Mapping}, where the phrase is aligned to the nearest canonical predicate (\eg, ``positioned right above'' $\rightarrow$ \texttt{above}, ``touches the body at the side'' $\rightarrow$ \texttt{attached-to}). Entities $i$ and $j$ are resolved to part indices using the PartVerse vocabulary or its synonyms. Ambiguities such as plural forms (``legs'') are resolved by mapping to all relevant slots, while singular references select a single part instance. Each triplet is interpreted as an assembly-level constraint that specifies how two parts are arranged.  
For example, $(\texttt{seat}, \texttt{legs}, \texttt{above})$ indicates that the seat is vertically supported by the legs,  
$(\texttt{handle}, \texttt{body}, \texttt{attached-to})$ encodes a functional attachment, and  
$(\texttt{wings}, \texttt{wings}, \texttt{symmetric-with})$ enforces bilateral symmetry.\looseness-1
\vspace{0.1cm}

\noindent\textbf{Validation.} To validate the generated functional and spatial relations in \datasetname, we adopt a two-stage protocol.  First, we perform \emph{geometric checks} on spatial triplets using the ground-truth part geometry. Each part mesh is loaded into Open3D, and its axis-aligned bounding box is computed directly from vertex coordinates. Predicate-specific inequalities are then applied to filter inconsistent or contradictory relations; triplets violating these constraints are flagged and removed. Second, we conduct a \emph{human audit} on the remaining triplets. In each run, we uniformly sample 200 triplets from the full dataset and manually verify their correctness using rendered multi-view images and part masks. We repeat this process 20 times to obtain a stable estimate of annotation quality across predicates and object types. Across all runs, spatial and functional triplets achieve an average correctness of 92\% and 88\%, respectively. During training, triplets are treated as \emph{relational signals}: they are embedded as relational semantic latents and aggregated through attention, allowing the model to down-weight inconsistent or noisy triplets.\looseness-1

\section{Implementation Details}\label{app:impldetails}
We train \modelnamenc in two stages using our relational \datasetname dataset introduced in Sec.~\ref{subsec:dataset}. In the first stage, we optimize the part latent with semantic synchronization under the DPL-RSL interaction framework. The diffusion backbone adopts a Transformer-based DiT architecture with cross-attention layers that enable joint reasoning across modalities and parts. To enhance the part-level representation, we fine-tune the VAE using PartVerse \cite{dong2025one} and PartNet \cite{mo2019partnet}. In the second stage, we fine-tune the full model jointly, including both part-level and object-level synchronization, with persistent relational semantic latents providing structural conditions throughout denoising. The training objective employs an SNR-weighted curriculum that gradually shifts the emphasis from low-level denoising toward high-level semantic alignment, progressively strengthening relational and structural coherence across parts. We use AdamW with a learning rate of $1\times10^{-4}$, cosine decay, and gradient clipping at 1.0. All experiments are conducted on four NVIDIA L40 GPUs. 
To ensure fair comparison, all models are evaluated on the same test split of the selected datasets.  
Baseline methods are evaluated using their official publicly available implementations, following the protocols recommended in their repositories. \looseness-1

\section{Additional Experiments}\label{app:more_results}

\begin{figure}
    \centering
    \includegraphics[width=0.99\linewidth]{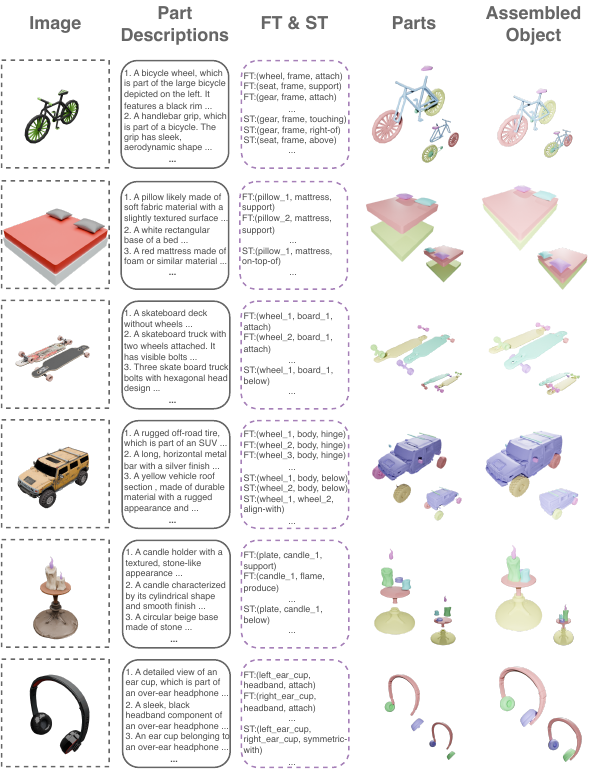}
    \vspace{-0.3cm}
    \captionof{figure}{\textbf{Qualitative Results.} Each column shows one stage of our generation process: the input image (optional), textual descriptions, canonicalized functional and spatial triplets (FT \& ST), \modelname-generated parts, and the final assembled object. As illustrated, \modelnamegradient enables high-quality part-aware 3D generation with no explicit grounding (\eg, bounding boxes).}
    \label{fig:qualitative_r2}
    \vspace{-0.3cm}
\end{figure}

\noindent\textbf{Qualitative Examples.} Figure~\ref{fig:qualitative_r2} illustrates the overall generation process, showing each stage of \modelnamenc{} from textual input to final 3D assembly. 
Given a textual input with optional image input, they are then enriched through canonicalized \textit{functional} and \textit{spatial triplets} (FT \& ST), which encode inter-part relations. Leveraging these structured representations, the model synthesizes high-quality parts and semantically coherent objects. As shown in the last two columns, \modelname successfully captures both individual parts and their global arrangement, enabling controllable and interpretable 3D generation without explicit geometric supervision or bounding-box guidance.
\vspace{0.2cm}

\begin{table*}[t!]
    \centering
    \caption{\textbf{Perceptual evaluation on part-level 3D object generation.}
    r-FID/r-KID are render-FID/render-KID from multi-view renderings; P-FID/P-KID computed in PointNet++~\cite{qi2017pointnet++} feature space. Highlighted \colorbox{CustomLightPurple}{best} and \colorbox{CustomLightLightPurple}{second best}.}
    \label{tab:quant-r4}
    \vspace{-0.3cm}
    \setlength{\tabcolsep}{3pt}
\resizebox{\linewidth}{!}{%
\begin{tabular}{l cccc cccc cccc cccc}
    \toprule
    \multirow{2}{*}{\textbf{Method}}
    & \multicolumn{4}{c}{\textbf{Objaverse}}
    & \multicolumn{4}{c}{\textbf{ShapeNet}}
    & \multicolumn{4}{c}{\textbf{ABO}}
    & \multicolumn{4}{c}{\textbf{PartRel3D}} \\
    \cmidrule(lr){2-5}\cmidrule(lr){6-9}\cmidrule(lr){10-13}\cmidrule(lr){14-17}
    & r-FID & r-KID & P-FID & P-KID
    & r-FID & r-KID & P-FID & P-KID
    & r-FID & r-KID & P-FID & P-KID
    & r-FID & r-KID & P-FID & P-KID \\
    \midrule
    Trellis
    & 5.487 & 2.100 & 0.231 & 1.300
    & 6.514 & 2.700 & 0.516 & 3.600
    & 5.924 & 3.100 & 0.448 & 2.300
    & 11.983 & 5.400 & 0.845 & 5.600 \\
    CLAY
    & 5.292 & 1.900 & 0.218 & 1.200
    & 6.328 & 2.400 & 0.500 & 3.400
    & 5.807 & 2.900 & 0.432 & 2.100
    & 11.761 & 5.200 & 0.822 & 5.400 \\
    HoloPart
    & \cellcolor{CustomLightLightPurple}4.924 & 1.800 & \cellcolor{CustomLightLightPurple}0.205 & 1.100
    & 5.871 & 2.200 & 0.473 & 3.200
    & 5.363 & \cellcolor{CustomLightLightPurple}2.400 & 0.402 & 1.900
    & \cellcolor{CustomLightLightPurple}10.947 & 4.800 & 0.793 & 5.000 \\
    PartCrafter
    & 5.015 & \cellcolor{CustomLightLightPurple}1.700 & 0.213 & \cellcolor{CustomLightLightPurple}1.000
    & \cellcolor{CustomLightLightPurple}5.539 & \cellcolor{CustomLightLightPurple}2.000 & \cellcolor{CustomLightLightPurple}0.451 & \cellcolor{CustomLightLightPurple}2.900
    & \cellcolor{CustomLightLightPurple}5.118 & \cellcolor{CustomLightLightPurple}2.400 & \cellcolor{CustomLightLightPurple}0.383 & \cellcolor{CustomLightLightPurple}1.800
    & 11.136 & \cellcolor{CustomLightLightPurple}4.500 & \cellcolor{CustomLightLightPurple}0.752 & \cellcolor{CustomLightLightPurple}4.700 \\
    \modelnamegradient
    & \cellcolor{CustomLightPurple}4.058 & \cellcolor{CustomLightPurple}1.200 & \cellcolor{CustomLightPurple}0.168 & \cellcolor{CustomLightPurple}0.900
    & \cellcolor{CustomLightPurple}4.974 & \cellcolor{CustomLightPurple}1.700 & \cellcolor{CustomLightPurple}0.413 & \cellcolor{CustomLightPurple}2.500
    & \cellcolor{CustomLightPurple}4.563 & \cellcolor{CustomLightPurple}2.000 & \cellcolor{CustomLightPurple}0.350 & \cellcolor{CustomLightPurple}1.500
    & \cellcolor{CustomLightPurple}9.784 & \cellcolor{CustomLightPurple}3.900 & \cellcolor{CustomLightPurple}0.692 & \cellcolor{CustomLightPurple}4.300 \\
    \bottomrule
\end{tabular}
}
\vspace{-0.2cm}
\end{table*}

\noindent\textbf{Perceptual Evaluation.} We report r-FID/KID and P-FID/P-KID separately in the \Cref{tab:quant-r4}. As shown, \modelnamenc{} achieves the best perceptual performance across all four datasets, with HoloPart and PartCrafter alternating as the strongest baselines depending on the metric. These results mirror the trends observed in the geometric evaluations, further confirming the advantages of our relational-aware generative framework.\looseness-1
\vspace{0.2cm}

\noindent\textbf{Condition-wise Analysis.} Table~\ref{tab:quant-r1} reports quantitative results under different conditioning setups. 
Among single-condition variants, text provides the strongest overall geometric and semantic performance. When combined with text, spatial triplets (ST) deliver the largest improvement over the text-only baseline. This confirms that language-grounded spatial relations provide strong geometric priors that guide assembly and alignment without requiring explicit 3D bounding-box supervision. In contrast, functional triplets (FT) alone perform less effectively, as their high-level semantics (\eg \emph{support}, \emph{attach}, \emph{hinge}) are linguistically abstract and do not directly constrain geometry. 
However, FT plays a complementary role by bridging the gap between textual intent and geometric structure. When combined with ST, it improves functional coherence across parts and stabilizes relational learning. Remarkably, the combined \texttt{Text+FT+ST} setting achieves performance that is competitive with and, across several metrics, nearly matches the full \texttt{Text+Image+FT+ST} configuration despite using no image input at all. These results show that relational triplets provide substantial supervision, demonstrating that structured linguistic relationships (functional + spatial) can encode much of the geometric and compositional information typically learned from visual cues. Finally, adding image guidance (\texttt{Text+Image+FT+ST}) produces the strongest overall performance, confirming that visual evidence and relational reasoning are synergistic.
\vspace{0.2cm}

\begin{table*}[t!]
\centering
\setlength{\tabcolsep}{4pt}
\setlength\extrarowheight{1pt}
\caption{\textbf{Quantitative evaluation with different input settings.}
We evaluate \modelname with different combinations of inputs. Among single-source inputs, text provides the strongest overall geometric and semantic performance, while functional and spatial triplets alone provide weaker constraints.
Combining text, image, and relational triplets yields the best or tied-best performance on most metrics, demonstrating the complementary benefits of language-grounded functional and spatial relations.}

\label{tab:quant-r1}
\vspace{-0.3cm}
\resizebox{\textwidth}{!}{
    \begin{tabular}{l c c c c c c c c}
    \toprule
    \textbf{Condition} 
    & CD$\downarrow$ 
    & EMD$\downarrow$ 
    & r-FID$\downarrow$ 
    & r-KID$\downarrow$ 
    & P-FID$\downarrow$ 
    & P-KID$\downarrow$ 
    & CLIP(I-T)$\uparrow$ 
    & ULIP-T$\uparrow$ \\
    \midrule
    
Image
& 0.174 & 1.272 & 8.573 & 3.000 & 1.065 & 2.000 & 0.235 & 0.155 \\

Text
& 0.167 & 1.264 & 8.091
& \cellcolor{CustomLightLightPurple}2.600
& 1.210 & 2.100 & 0.239 & 0.158 \\

Functional Triplets (FT)
& 0.180 & 1.348 & 9.214 & 3.100 & 1.324 & 2.300 & 0.201 & 0.141 \\

Spatial Triplets (ST)
& 0.179 & 1.321 & 8.932 & 2.900 & 1.278 & 2.200 & 0.214 & 0.147 \\

\midrule

Text+FT
& 0.150 & 0.821 & 7.032 & 2.700 & 0.948 & 1.600 & 0.241 & 0.164 \\

Text+ST
& 0.112 & 0.298 & 6.842
& \cellcolor{CustomLightLightPurple}2.600
& 0.782 & 1.600
& \cellcolor{CustomLightLightPurple}0.245
& 0.169 \\

Text+FT+ST
& \cellcolor{CustomLightLightPurple}0.085
& \cellcolor{CustomLightLightPurple}0.161
& \cellcolor{CustomLightLightPurple}5.708
& \cellcolor{CustomLightPurple}1.800
& \cellcolor{CustomLightPurple}0.701
& \cellcolor{CustomLightPurple}1.100
& \cellcolor{CustomLightLightPurple}0.245
& \cellcolor{CustomLightLightPurple}0.174 \\

\textbf{Text+Image+FT+ST}
& \cellcolor{CustomLightPurple}0.080
& \cellcolor{CustomLightPurple}0.147
& \cellcolor{CustomLightPurple}5.432
& \cellcolor{CustomLightPurple}1.800
& \cellcolor{CustomLightLightPurple}0.725
& \cellcolor{CustomLightPurple}1.100
& \cellcolor{CustomLightPurple}0.251
& \cellcolor{CustomLightPurple}0.176 \\

\bottomrule
    \end{tabular}
}
\vspace{-0.2cm}
\end{table*}

\begin{wraptable}{r}{0.65\linewidth}
\vspace{-1.1cm}
\centering
\caption{\textbf{Quantitative evaluation on part-level 3D generation.}
We compare methods that explicitly generate part meshes on part-annotated datasets.
\colorbox{CustomLightPurple}{Best} and \colorbox{CustomLightLightPurple}{second-best} results are highlighted.}
\label{tab:quant_part_level}
\setlength{\tabcolsep}{4pt}
\resizebox{\linewidth}{!}{%
\begin{tabular}{l c c c  c c c}
    \toprule
    \multirow{2}{*}{\textbf{Method}} 
    & \multicolumn{3}{c}{\textbf{ShapeNet}} 
    & \multicolumn{3}{c}{\textbf{PartRel3D}} \\
    \cmidrule(lr){2-4} \cmidrule(lr){5-7}
    & CD$\downarrow$ & EMD$\downarrow$ & F-Score$\uparrow$
    & CD$\downarrow$ & EMD$\downarrow$ & F-Score$\uparrow$ \\
    \hline
    HoloPart
    & 0.162 & 0.625 & \cellcolor{CustomLightLightPurple}0.758
    & 0.153 & \cellcolor{CustomLightLightPurple}0.598 & 0.741 \\
    PartCrafter
    & \cellcolor{CustomLightLightPurple}0.141 & \cellcolor{CustomLightLightPurple}0.603 & 0.732
    & \cellcolor{CustomLightLightPurple}0.137 & 0.612 & \cellcolor{CustomLightLightPurple}0.756 \\
    \modelnamegradient
    & \cellcolor{CustomLightPurple}0.088 & \cellcolor{CustomLightPurple}0.451 & \cellcolor{CustomLightPurple}0.863
    & \cellcolor{CustomLightPurple}0.081 & \cellcolor{CustomLightPurple}0.438 & \cellcolor{CustomLightPurple}0.772 \\
    \bottomrule
\end{tabular}}
\vspace{-0.8cm}
\end{wraptable}

\noindent\textbf{Part-Level Generation.}
\Cref{tab:quant_part_level} complements our object-level evaluation by measuring reconstruction fidelity on individual generated parts (CD and EMD). 
We additionally report the F-score at threshold 0.005 by computing precision/recall between sampled points from the generated and ground-truth part surfaces. Across part-annotated datasets, \modelnamenc consistently achieves the best per-part geometry quality, indicating that its gains are not only due to improved global assembly but also stronger generation of each component. In particular, the improvements in F-score show that \modelnamenc recovers more accurate part surfaces rather than merely reducing average distance metrics, confirming that the proposed DPL-RSL synchronization benefits fine-grained part geometry generation in addition to overall object coherence.
\vspace{0.2cm}

\definecolor{DeltaColor}{RGB}{180,70,70}
\newcommand{\deltaval}[1]{\textcolor{DeltaColor}{($\Delta$ #1)}}

\begin{table}[t!]
    \vspace{-0.3cm}
    \centering
    \caption{\textbf{Relation parsing robustness at inference.} We compare three inference conditions for constructing relational triplets at test time: (1) with an additional VLM (Qwen2.5-VL or GPT-5), (2) oracle relations with ground-truth triplets, and (3) without VLMs (prompt-only conditioning). For prompt-only, we report the gap \deltaval{} to the best-performing variant for each metric. Highlighted \colorbox{CustomLightPurple}{best} and \colorbox{CustomLightLightPurple}{second-best}.}
    \label{tab:vlm_reliance}
        \vspace{-0.3cm}
            \setlength{\tabcolsep}{5pt}
    \resizebox{\linewidth}{!}{
        \begin{tabular}{l c c c c}
        \toprule
        \textbf{Method} & {r-FID}$\downarrow$ & {CD}$\downarrow$ & {ULIP-T}$\uparrow$ & {IoU}$\downarrow$ \\
        \midrule
        HoloPart & 10.942 & 0.334 & 0.113 & 0.723 \\
        PartCrafter & 11.134 & 0.312 & 0.109 & 0.717 \\
        \modelnamegradient + Qwen2.5-VL 
        & \cellcolor{CustomLightLightPurple}9.701 
        & 0.101 
        & \cellcolor{CustomLightPurple}0.161 
        & 0.491 \\
        \modelnamegradient + GPT-5 
        & 9.744 
        & \cellcolor{CustomLightPurple}0.097 
        & \cellcolor{CustomLightLightPurple}0.153 
        & \cellcolor{CustomLightPurple}0.469 \\
        \modelnamegradient + Oracle 
        & \cellcolor{CustomLightPurple}9.684 
        & 0.101 
        & \cellcolor{CustomLightPurple}0.161 
        & \cellcolor{CustomLightLightPurple}0.471 \\
        \modelnamegradient            
        & 9.783 \deltaval{0.099} 
        & \cellcolor{CustomLightLightPurple}0.099 \deltaval{0.002} 
        & \cellcolor{CustomLightLightPurple}0.153 \deltaval{0.008} 
        & 0.474 \deltaval{0.005} \\
        \bottomrule
        \end{tabular}
    }
\end{table}
\noindent \textbf{Robustness to Relation Parsing.}
\label{subsec:vlm_robustness}
A key question is whether \modelnamenc{} depends on a particular relational parser at inference time to supplement FT and ST, or whether the generative model itself has already internalized the relational structure knowledge. To study this, we evaluate three inference settings: (i) \textbf{VLM-parsed relations}, with two parser variants: the same VLM used for dataset construction (Qwen2.5-VL) and a stronger external parser (GPT-5). (ii) \textbf{prompt-only conditioning} without explicit relation parsing, and (iii) \textbf{oracle relations} using ground-truth triplets. As shown in \Cref{tab:vlm_reliance}, prompt-only inference remains competitive, indicating that the model internalizes substantial part-level and assembly priors during training. The small gap between Qwen2.5-VL and GPT-5 further suggests that the gains come from the RSL mechanism rather than parser-specific artifacts.
\vspace{0.2cm}

\begin{wraptable}{r}{0.6\linewidth}
\vspace{-1.2cm}
\setlength{\tabcolsep}{3pt}
\centering
\caption{\textbf{Ablation on the number of local RSL tokens $K_m$.} We vary the number of semantic controller tokens used for local relational-semantic guidance. Highlighted \colorbox{CustomLightPurple}{best} and \colorbox{CustomLightLightPurple}{second-best} results.}
\setlength{\tabcolsep}{5pt}
\resizebox{\linewidth}{!}{%
\begin{tabular}{c c c c c c c}
    \toprule
    \textbf{$K_m$} & CD$\downarrow$ & EMD$\downarrow$ & IoU$\downarrow$ & CLIP(N-T)$\uparrow$ & CLIP(I-T)$\uparrow$ & ULIP-T$\uparrow$ \\
    \midrule
    8  & 0.091 & 0.438 & 0.302 & 0.171 & 0.193 & 0.147 \\
    \cellcolor{CustomLightPurple}16 & \cellcolor{CustomLightPurple}0.084 & \cellcolor{CustomLightPurple}0.421 & \cellcolor{CustomLightPurple}0.286 & \cellcolor{CustomLightPurple}0.179 & \cellcolor{CustomLightPurple}0.200 & \cellcolor{CustomLightPurple}0.153 \\
    32 & \cellcolor{CustomLightLightPurple}0.085 & \cellcolor{CustomLightLightPurple}0.423 & \cellcolor{CustomLightLightPurple}0.301 & 0.178 & \cellcolor{CustomLightLightPurple}0.199 & \cellcolor{CustomLightLightPurple}0.152 \\
    64 & 0.087 & 0.425 & \cellcolor{CustomLightLightPurple}0.301 & \cellcolor{CustomLightLightPurple}0.177 & 0.189 & \cellcolor{CustomLightPurple}0.153 \\

    \bottomrule
\end{tabular}}
\label{tab:rsl_k}
\vspace{-0.8cm}
\end{wraptable}\noindent\textbf{Number of Local RSL Tokens.}
RSLs act as semantic controllers, and their count $K_m$ reflects the number of meaningful part-level attributes or relations. In \datasetname, most objects contain roughly 10--30 such cues, so the token budget naturally remains small. We therefore evaluate $K_m \!\in\! \{8,16,32,64\}$, a range that covers typical semantic density while keeping diffusion attention efficient.  As shown in Table \ref{tab:rsl_k}, performance stabilizes once $K_m \ge 16$, indicating that only a modest number of semantic tokens is needed for strong guidance. We set $K_m = 16$ as the default in all experiments.
\vspace{0.2cm}

\noindent\textbf{Generalization Beyond Clean Part Decompositions.}
\label{subsec:novel_parts}
A key concern for part-based generators is reliance on clean, taxonomy-consistent part decompositions. To quantify robustness beyond the most common training configurations, we construct two out-of-distribution (OOD) evaluation splits that probe novel part and novel relation generalization.
(i) \textbf{OOD-parts (rare-part split):} we compute the training-set frequency of each part label (object-level occurrence) and define \emph{rare} parts as those in the tail of this distribution, with a minimum-count filter (at least 2) to avoid noisy labels; the OOD-parts split includes all test objects that contain at least one rare part label.
(ii) \textbf{OOD-rel (novel-relation split):} we hold out a subset of relation predicates during training by removing all triplets whose predicate $\rho$ belongs to a held-out set, and evaluate on test samples that include at least one held-out predicate.
We report the same fidelity, alignment, and structure metrics as in the main evaluation (r-FID, CD, ULIP-T, and IoU as a part-independence measure).
As shown in \Cref{tab:novel_parts}, all methods degrade under OOD shifts, but \modelname exhibits smaller performance drops than prior part-based baselines: for example, under OOD-rel, PartCrafter increases from 11.134 to 12.583 in r-FID ($\Delta$ 1.449), while \modelname increases from 9.783 to 10.631 ($\Delta$ 0.848). Moreover, \modelname maintains strong text-shape alignment under both splits (ULIP-T drops by only $\Delta$ 0.012-0.014), indicating that the learned relational priors generalize beyond the dominant training taxonomy and support coherent assembly even when parts or relations are less common.\looseness-1

\begin{table}[t]
\centering

\caption{\textbf{Robustness to rare parts and held-out relations.} 
We evaluate in-distribution (ID) test data and two out-of-distribution (OOD) splits: OOD-parts (tail part labels) and OOD-rel (held-out relation predicates). 
We report absolute scores and the change relative to ID in parentheses \deltaval{}.}
\label{tab:novel_parts}
\vspace{-1em}
\setlength{\tabcolsep}{6pt}
\resizebox{\columnwidth}{!}{
\begin{tabular}{l l c c c c}
\toprule
\textbf{Method} & \textbf{Split} & {r-FID}$\downarrow$ & {CD}$\downarrow$ & {ULIP-T}$\uparrow$ & {IoU}$\downarrow$ \\
\midrule
\multirow{3}{*}{HoloPart} 
& ID  & 10.942 & 0.334 & 0.113 & 0.723 \\
& OOD-parts & 12.318 \deltaval{1.376} & 0.392 \deltaval{0.058} & 0.101 \deltaval{0.012} & 0.781 \deltaval{0.058} \\
& OOD-rel   & 12.701 \deltaval{1.759} & 0.408 \deltaval{0.074} & 0.098 \deltaval{0.015} & 0.797 \deltaval{0.074} \\
\midrule
\multirow{3}{*}{PartCrafter} 
& ID  & 11.134 & 0.312 & 0.109 & 0.717 \\
& OOD-parts & 12.206 \deltaval{1.072} & 0.358 \deltaval{0.046} & 0.097 \deltaval{0.012} & 0.759 \deltaval{0.042} \\
& OOD-rel   & 12.583 \deltaval{1.449} & 0.371 \deltaval{0.059} & 0.094 \deltaval{0.015} & 0.771 \deltaval{0.054} \\
\midrule
\multirow{3}{*}{\modelnamegradient}
& ID  & 9.783 & 0.099 & 0.153 & 0.474 \\
& OOD-parts & 10.412 \deltaval{0.629} & 0.113 \deltaval{0.014} & 0.141 \deltaval{0.012} & 0.506 \deltaval{0.032} \\
& OOD-rel   & 10.631 \deltaval{0.848} & 0.118 \deltaval{0.019} & 0.139 \deltaval{0.014} & 0.519 \deltaval{0.045} \\
\bottomrule
\end{tabular}}
\end{table}

\section{Applications}
\label{app:applications}
\subsection{Mini-scene Generation}
\begin{figure}[t!]
    \centering
    \includegraphics[width=0.99\linewidth]{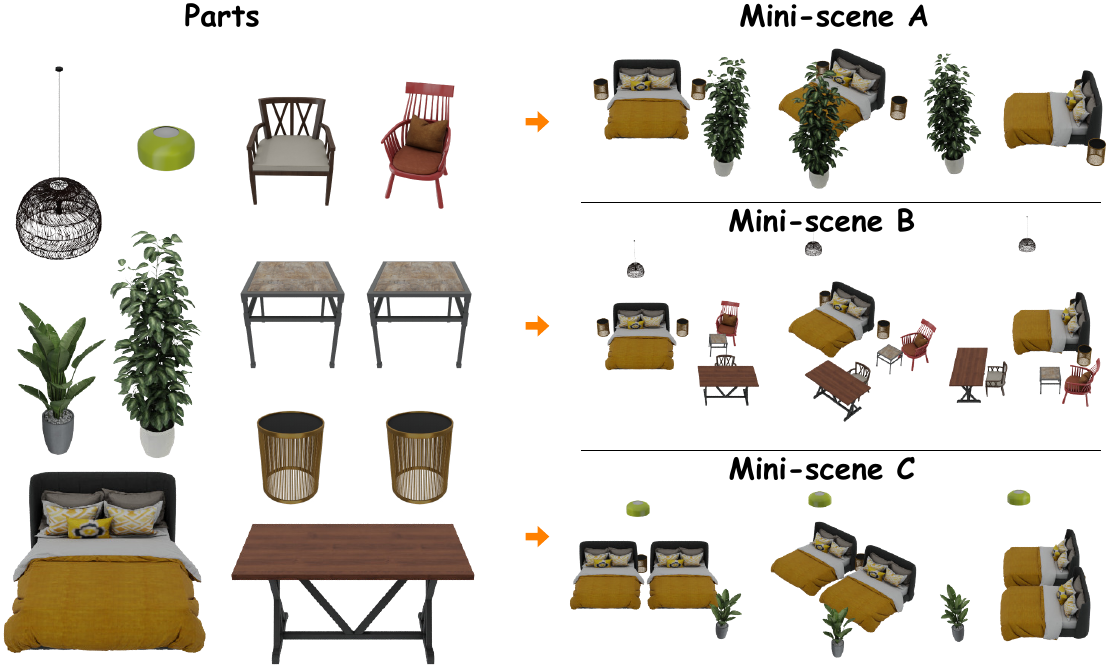}
    \vspace{-0.3cm}
    \caption{\textbf{Additional Mini-scene Generation Results.}}
    \label{fig:more_miniscene}
\end{figure}

In this task, \modelname generates a coherent multi-object arrangement (a small scene) directly from a text prompt that describes several semantically related objects and their spatial relations. During generation, we treat each object as a macro-part, represented by its aggregated DPLs $(\mathbf{L}^{\text{3D}}_i, \mathbf{L}^{\text{2D}}_i)$ and a relational graph derived from scene-level captions. These scene graphs are constructed using the same canonicalization procedure, producing inter-object triplets $(o_i, o_j, \rho)$ that describe spatial and functional relations. The resulting scene-level semantic tokens $\mathbf{S}^{\text{scene}}$ guide object placement through cross-object attention, ensuring spatial consistency while preserving each object’s internal structure. To synthesize a complete scene, objects are first sampled independently and then jointly refined by re-synchronizing their DPLs under $\mathbf{S}^{\text{scene}}$. Quantitatively, \Cref{tab:scene_gen} shows that \modelname improves both geometric fidelity and compositional consistency over prior methods, achieving lower CD and higher F-score. \Cref{fig:more_miniscene} further demonstrates that this process yields diverse, coherent mini-scenes.\looseness-1
\begin{table}[t]
\centering
\setlength{\tabcolsep}{6pt}
\caption{\textbf{Evaluation on 3D Object-Composed Scene Generation.}
We report results on \textbf{3D-Front} and an \textbf{Occluded} subset.
We report CD$\downarrow$, F-score$\uparrow$, and IoU$\downarrow$, together with inference runtime$\downarrow$ per scene. Highlighted \colorbox{CustomLightPurple}{best} and \colorbox{CustomLightLightPurple}{second-best} results.}
\label{tab:scene_gen}
\vspace{-0.3cm}
\resizebox{\columnwidth}{!}{
\begin{tabular}{l  c c c  c c c  c}
\toprule
\multirow{2}{*}{\textbf{3D Scene Generation}} 
& \multicolumn{3}{c}{\textbf{3D-Front}} 
& \multicolumn{3}{c}{\textbf{3D-Front (Occluded)}} 
& \multirow{2}{*}{\textbf{Run Time}$\downarrow$} \\
\cmidrule(lr){2-4} \cmidrule(lr){5-7} 
& {CD}$\downarrow$ & {F-Score}$\uparrow$ & {IoU}$\downarrow$
& {CD}$\downarrow$ & {F-Score}$\uparrow$ & {IoU}$\downarrow$
& \\
\midrule
MIDI \cite{huang2025midi} 
& 0.1602 & 0.7931 & \cellcolor{CustomLightLightPurple}0.0013
& 0.2591 & 0.6618 & \cellcolor{CustomLightLightPurple}0.0020
& 80s \\
PartCrafter 
& \cellcolor{CustomLightLightPurple}0.1528 & \cellcolor{CustomLightLightPurple}0.8085 & 0.0016
& \cellcolor{CustomLightLightPurple}0.2387 & \cellcolor{CustomLightLightPurple}0.7042 & 0.0022
& \cellcolor{CustomLightLightPurple}42s \\
\modelnamegradient 
& \cellcolor{CustomLightPurple}0.1495 & \cellcolor{CustomLightPurple}0.8146 & \cellcolor{CustomLightPurple}0.0012
& \cellcolor{CustomLightPurple}0.2321 & \cellcolor{CustomLightPurple}0.7128 & \cellcolor{CustomLightPurple}0.0019
& \cellcolor{CustomLightPurple}40s \\
\bottomrule
\end{tabular}
}
\end{table}

\subsection{Articulated Object Generation}
To model articulation, we first construct paired configurations of the same object representing opposite canonical poses. Following~\cite{gao2024partgs,yu2026part}, we estimate per-part transformations $\mathcal{T}_i\!=\!(\mathbf{R}_i, \mathbf{t}_i)$ by aligning the corresponding parts across the two states. 
Each part is first identified through its learned part-identifier embedding
$\mathbf{e}_i$, which is held fixed during articulation estimation, and the transformation parameters are derived via rigid motion fitting between the part’s geometry in the two poses. This yields a compact articulation field $\{\mathcal{T}_i\}_{i=1}^N$ describing how each part moves relative to its canonical configuration.
Once transformations are obtained, we reconstruct articulated motion by applying $\mathcal{T}_i$ to the canonical part meshes and reassembling the original objects. The resulting articulated objects maintain structural integrity across states and preserve semantic consistency through the persistent part embeddings. This setup allows us to visualize or simulate motion between poses without any re-optimization or diffusion-based retraining. As illustrated in \Cref{fig:more_arti}, our relationally grounded model naturally produces articulated 3D assets that preserve structural consistency across different motion states.\looseness-1

\subsection{Part Editing}
\begin{figure}[htb!]
\centering

    \begin{minipage}[t]{0.49\linewidth}
      \centering
      \raisebox{0pt}[\height][0pt]{%
        \includegraphics[height=0.80\textheight]{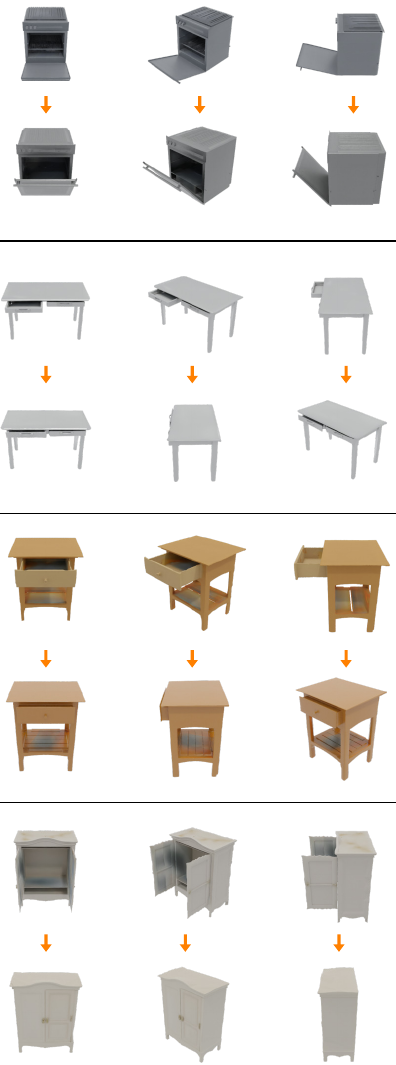}%
      }
      \vspace{-0.25cm}
      \caption{\textbf{Part Articulation Results.} \modelname naturally produces articulated 3D assets that preserve structural consistency across different motion states.}
      \label{fig:more_arti}
    \end{minipage}\hfill
    \begin{minipage}[t]{0.49\linewidth}
      \centering
      \raisebox{0pt}[\height][0pt]{%
        \includegraphics[height=0.80\textheight]{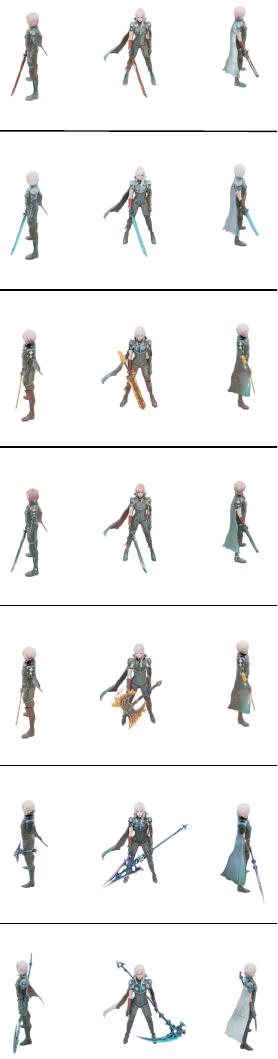}%
      }
      \vspace{-0.25cm}
      \caption{\textbf{Additional Part Editing Results.} The semantically grounded training of \modelnamenc allows diverse yet consistent part-level 3D asset generation.}
      \label{fig:more_part_editing}
    \end{minipage}

\end{figure}

To edit a specific part, we isolate its DPLs using the part identifier and freeze all non-target slots, keeping the global relational context $\mathbf{S}^{\text{glb}}$ fixed. We then perform localized re-denoising via partial DDIM inversion: the object is inverted to an intermediate noise level $\tau$, and only the target part’s 3D and 2D latents are optimized. Afterward, the updated DPLs are decoded and briefly re-synchronized with the rest of the object to ensure structural coherence. More results on part editing are available in \Cref{fig:more_part_editing}.

\section{Broader Impacts}
The ability to generate, compose, and edit 3D objects at the part level has broad implications across robotics, simulation, virtual content creation, and digital twin systems. \modelnamenc contributes to this space by offering a semantically grounded framework that produces structurally coherent, fine-grained 3D assets directly from language. This capability can enhance how embodied agents reason about objects, support richer interaction models in simulation, and accelerate the creation of editable assets for entertainment, industrial design, and education. In practical settings, such compositional generation can reduce the cost and expertise barrier for producing accurate and customizable 3D models, benefiting designers, animators, and researchers who rely on physically meaningful structures. 

At the same time, generative systems of this kind carry risks, including potential privacy concerns when reconstructing real-world objects, intellectual property considerations when producing stylized assets, and misuse in synthetic media pipelines. Although \modelnamenc is intended for research and educational use, we encourage responsible deployment practices that respect consent, attribution, and content integrity. Its modular and transparent design does not eliminate the need for careful governance. Deployment should still follow best practices around provenance, data consent, and domain-specific usage guidelines. Overall, we believe the benefits of controllable, semantically structured 3D generation outweigh the risks when accompanied by appropriate oversight and ethical use.

\end{document}